\documentclass[11pt]{article}
\pdfoutput=1

\usepackage[preprint]{acl}

\usepackage{times}
\usepackage{latexsym}

\usepackage[T1]{fontenc}

\usepackage[utf8]{inputenc}

\usepackage{microtype}

\usepackage{inconsolata}

\usepackage{graphicx}
\usepackage{soulutf8}
\usepackage{booktabs}
\usepackage{tabularx}
\usepackage{colortbl}
\usepackage{rotating}
\usepackage{amsmath} 
\usepackage{subcaption}
\usepackage{caption}
\usepackage{placeins}
\usepackage{xspace}
\usepackage{multirow}
\usepackage{enumitem}
\usepackage{fontawesome5} 
\usepackage{epigraph}

\definecolor{peach}           {HTML}{f5e2d5}  %
\definecolor{palCream}        {HTML}{fbf5ef}  %
\definecolor{palWarmTan}      {HTML}{b09a7a}  %
\definecolor{palDarkCharcoal} {HTML}{3d3530}  %
\definecolor{palTerracotta}   {HTML}{9c4a2f}  %
\definecolor{palDeepTeal}     {HTML}{2f6b70}  %

\colorlet{todoback}  {palTerracotta!8}
\colorlet{todoborder}{palTerracotta!55}
\colorlet{todoline}  {palWarmTan}

\usepackage[
  textsize=scriptsize,
  textwidth=2cm,
  backgroundcolor=todoback,
  bordercolor=todoborder,
  linecolor=todoline,
  tickmarkheight=3pt,
  disable,
]{todonotes}

\newcommand{\todoin}[1]{}
\newcommand{\dw}[1]{}

\newcolumntype{Y}[1]{>{\raggedright\arraybackslash\hsize=#1\hsize}X}

\newcommand{\humanonly}{\texttt{Human-Only}\xspace}
\newcommand{\humangen}{\texttt{Human+\allowbreak Generated}\xspace}

\newcommand{\Dpos}{\ensuremath{D_{+}}}
\newcommand{\Dneg}{\ensuremath{D_{-}}}

\usepackage{listings}
\usepackage{listingsutf8}  %
\usepackage[most]{tcolorbox}
\tcbuselibrary{skins,breakable,listings}
\usepackage{float}

\usepackage{newfloat}
\DeclareFloatingEnvironment[
  fileext=lop,
  listname={List of Prompts},
  name={Prompt},
  placement=tbp,
]{Prompt}

\colorlet{promptbg}{palCream}            %
\colorlet{promptframe}{palWarmTan}       %
\colorlet{promptaccent}{palDarkCharcoal} %
\colorlet{promptslot}{palTerracotta}     %
\colorlet{promptins}{palDeepTeal}        %

\newcommand{\promptfontsize}{\small}
\newcommand{\promptfontsmall}{\fontsize{8}{9.2}\selectfont}
\newcommand{\promptfonttiny}{\fontsize{7.5}{8.6}\selectfont}

\lstdefinestyle{promptstyle}{
  basicstyle=\promptfontsize\ttfamily,
  backgroundcolor=\color{promptbg},
  inputencoding=utf8/latin1,
  breaklines=true,
  breakatwhitespace=false,
  keepspaces=true,
  numbers=none,
  showspaces=false,
  showstringspaces=false,
  showtabs=false,
  aboveskip=0pt,
  belowskip=0pt,
  xleftmargin=0pt,
  upquote=true,
}

\lstdefinestyle{promptskeletonstyle}{
  style=promptstyle,
  moredelim=[is][\color{promptslot}\bfseries]{<<}{>>},
}

\lstdefinestyle{promptdiffstyle}{
  style=promptskeletonstyle,
  moredelim=[is][\color{promptins}\bfseries]{[[}{]]},
}

\tcbset{
  promptskin/.style={
    enhanced,
    width=\linewidth,
    colback=promptbg,
    colframe=promptframe,
    boxrule=0.5pt,
    arc=3mm,
    outer arc=3mm,
    left=5mm, right=5mm, top=4mm, bottom=4mm,
    shadow={1.5mm}{-1.5mm}{0mm}{black!8},
    attach boxed title to top left={yshift=-3mm, xshift=5mm},
    boxed title style={
      colback=white,
      colframe=promptframe,
      arc=2mm, outer arc=2mm,
      boxrule=0.5pt,
      left=2mm, right=2mm, top=1mm, bottom=1mm,
    },
    fonttitle=\normalsize,
    coltitle=promptaccent,
  },
}

\tcbset{promptbreak/.style={breakable}}

\newtcblisting{promptbox}[2][]{%
  promptskin,
  promptbreak,
  title={#2},
  listing only,
  listing options={style=promptstyle},
  #1,
}

\newcommand{\promptfile}[3][]{%
  \begin{tcolorbox}[promptskin, promptbreak, title={#2}, #1]%
    \lstinputlisting[style=promptstyle]{#3}%
  \end{tcolorbox}%
}

\newcommand{\promptskeletonfile}[3][]{%
  \begin{tcolorbox}[promptskin, promptbreak, title={#2}, #1]%
    \lstinputlisting[style=promptskeletonstyle]{#3}%
  \end{tcolorbox}%
}

\newcommand{\promptdifffile}[3][]{%
  \begin{tcolorbox}[promptskin, promptbreak, title={#2}, #1]%
    \lstinputlisting[style=promptdiffstyle]{#3}%
  \end{tcolorbox}%
}

\newcommand{\slot}[1]{{\ttfamily\bfseries\color{promptslot}#1}}

\makeatletter
\newcommand{\prompt@short}{}
\newcommand{\promptshortcaption}[1]{\def\prompt@short{#1}}

\newcommand{\prompt@begin}[2]{%
  \def\prompt@cap{#1}%
  \def\prompt@lab{#2}%
  \def\prompt@short{}%
  \tcbset{promptbreak/.style={unbreakable}}%
  \centering
}
\newcommand{\prompt@end}{%
  \par\vspace{2pt}%
  \ifx\prompt@short\@empty
    \caption{\prompt@cap}%
  \else
    \expandafter\caption\expandafter[\prompt@short]{\prompt@cap}%
  \fi
  \label{\prompt@lab}%
}

\NewDocumentEnvironment{promptfigure}{O{tbp} m m}
  {\begin{Prompt}[#1]\prompt@begin{#2}{#3}}
  {\prompt@end\end{Prompt}}

\NewDocumentEnvironment{promptfigure*}{O{tp} m m}
  {\begin{Prompt*}[#1]\prompt@begin{#2}{#3}}
  {\prompt@end\end{Prompt*}}
\makeatother

\title{Old Ideas, Novel Problems:\\The Instability of LLM-Based Novelty Evaluation}

\newcommand{\codeurl}{https://github.com/allenai/novelty-eval}
\newcommand{\dataurl}{https://huggingface.co/datasets/allenai/novelty-judge-bench}
\newtcbox{\linkpill}{on line, enhanced, colback=palCream, colframe=palWarmTan,
  boxrule=0.4pt, arc=2.2mm, left=5pt, right=5pt, top=2.5pt, bottom=2.5pt,
  boxsep=0pt, fontupper=\small\sffamily\color{palDarkCharcoal}}

\newcommand{\paperlink}[3]{\href{#1}{\linkpill{#2\hspace{0.4em}#3}}}
\newcommand{\paperlinks}{%
  \normalfont\normalsize
  \paperlink{\codeurl}{\faGithub}{Code}\hspace{0.8em}%
  \paperlink{\dataurl}{\hficon}{Data}}

\makeatletter
\ifacl@anonymize
  \renewcommand{\codeurl}{https://anonymous.4open.science/r/novelty-eval-DDB5/README.md}
  \renewcommand{\dataurl}{https://anonymous.4open.science/r/novelty-eval-DDB5/data/README.md}
  \renewcommand{\paperlinks}{%
    \normalfont\normalsize
    \paperlink{\codeurl}{\faCode}{Code}\hspace{0.8em}%
    \paperlink{\dataurl}{\faDatabase}{Data}}
  \renewcommand\outauthor{%
    \begin{tabular}[t]{c}
      \bfseries Anonymous ACL submission \\[1.2ex]
      \paperlinks
    \end{tabular}}
\fi
\makeatother

\author{
  \textbf{Noy Sternlicht\textsuperscript{1,2}},
  \textbf{Simra Shahid\textsuperscript{3}},
  \textbf{Peter Jansen\textsuperscript{2,4}},
\\
  \textbf{Daniel S. Weld\textsuperscript{2,5}},
  \textbf{Pao Siangliulue\textsuperscript{2}},
  \textbf{Tom Hope\textsuperscript{1,2}}
\\
\\
  \textsuperscript{1}Hebrew University of Jerusalem,
  \textsuperscript{2}Allen Institute for AI,
  \textsuperscript{3}Microsoft,
\\
  \textsuperscript{4}University of Arizona,
  \textsuperscript{5}University of Washington
\\
\\
  \paperlinks}

\begin{document}
\maketitle

\addtocontents{toc}{\protect\setcounter{tocdepth}{-5}}
\makeatletter
\immediate\write\@auxout{\string\@writefile{lof}{\string\setcounter{tocdepth}{-5}}}
\immediate\write\@auxout{\string\@writefile{lot}{\string\setcounter{tocdepth}{-5}}}
\makeatother

\begin{abstract}
AI-based 
ideation systems are often evaluated on the novelty of
their ideas, and that judgment is
increasingly delegated to large language models. Such judges are typically built ad hoc and validated, if at all, on human-authored papers rather than on the generated ideas they are meant to score. So, how do novelty judges perform? Not well. 
We present a systematic controlled study of novelty evaluation design choices. 
We build an evaluation set automatically,
mining OpenReview for passages where reviewers explicitly affirm or dispute a paper's originality and keeping only submissions with unanimous agreement at the extremes of their research area; we pair these with ideas from a vanilla LLM generator. 
Across six judges, 
we find that 
seemingly small prompt design choices have large consequences;
e.g., simply telling the judge that reviewers found one idea novel and the other not can change its verdict on more than half of the identical idea pairs it is shown, shifting pairwise accuracy by over $50$ points and occasionally pushing it below chance. The same change helps one judge and hurts another. 
Retrieval and larger reasoning budgets help little, and two purpose-built novelty evaluators are outperformed by our cheapest prompted baseline.  
These results raise questions about reported novelty gains of automated ideation systems, 
and call for robust novelty evaluation methods.

\end{abstract}

\section{Introduction}

Recent years have seen a proliferation of systems that generate research ideas and propose new directions. A central criterion for judging such systems is \emph{novelty}: whether 
the ideas they produce contribute
a meaningful conceptual advance. Many works therefore automate this assessment, using it both to evaluate the ideas a system outputs and as an internal component of the ideation pipeline, letting agents filter and refine a pool of candidate ideas \cite{wang2024scimon,Yamada2025TheAS}.

\begin{figure}[t]
    \centering
    \includegraphics[width=0.9\columnwidth]{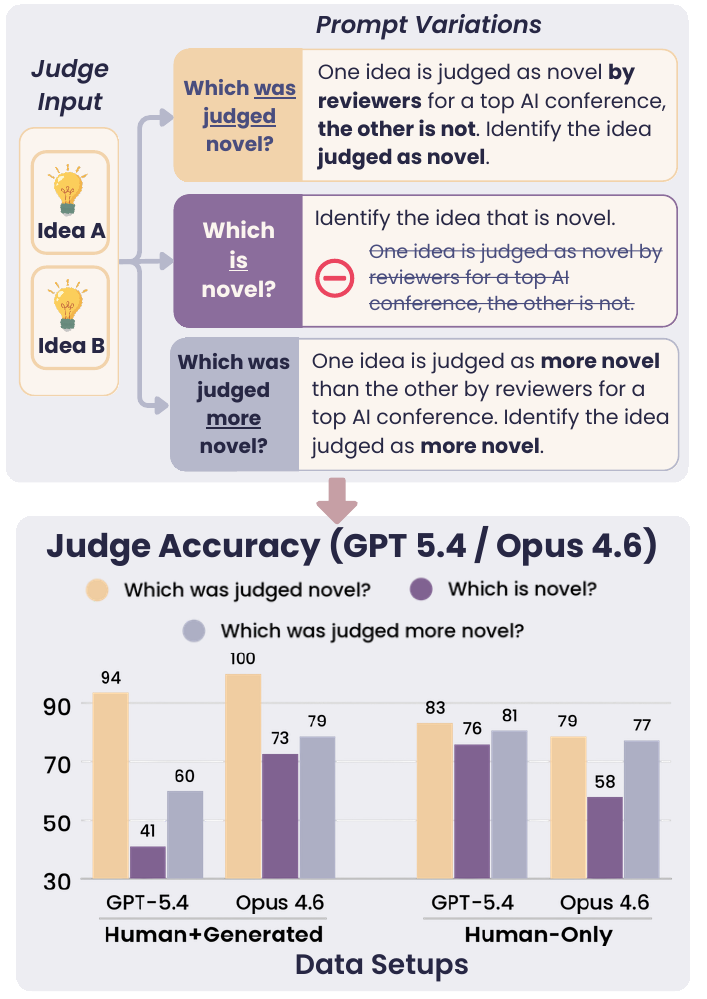}
    \caption{
    We run {a controlled study} of novelty judges.
    Small prompt changes swing accuracy on both human-authored (\humanonly) and LLM-generated (\humangen) ideas, more so on generated ones and unevenly across judges.
    }\label{fig:teaser}
\end{figure}

To date, there is no standard approach for novelty assessment, and papers suggesting a new ideation system typically ship their own ad hoc LLM judge \cite{radensky2026humanllmcompoundscientificideation,Gottweis2025TowardsAA}. Such judges are built around the system at hand, and might assume certain priors (e.g., a particular idea format). Their robustness, and the design choices behind them (such as their instructions, or whether they are allowed to access prior work), are rarely tested or justified. Validation rests on small annotated sets \cite{radensky2026humanllmcompoundscientificideation}, outdated data that predates LLMs' knowledge cutoffs \cite{shen2026unlockingllmcreativityscience}, or human-authored papers \cite{Si2024CanLG}, a different distribution from the generated ideas the judge is actually applied to \cite{chen2026measuringgaphumanllm}.

We present a systematic, extensive study of LLM-based novelty evaluation. Through controlled experiments, we vary the design choices practitioners make: which LLM is used as a judge and how the judge is prompted, whether evaluation is \emph{pointwise} (is this idea novel?) or \emph{pairwise} (which of these two is more novel?), its reasoning budget, whether it may retrieve related work, and the format of the judged ideas. We evaluate judges on human-authored ideas, and explore how their behavior changes when LLM-generated ideas are judged alongside them. We also benchmark off-the-shelf LLMs against specialized novelty judges \cite{Yamada2025TheAS,Shahid2025LiteratureGroundedNA}.

\begin{figure*}[t]
    \centering
    \includegraphics[width=\textwidth]{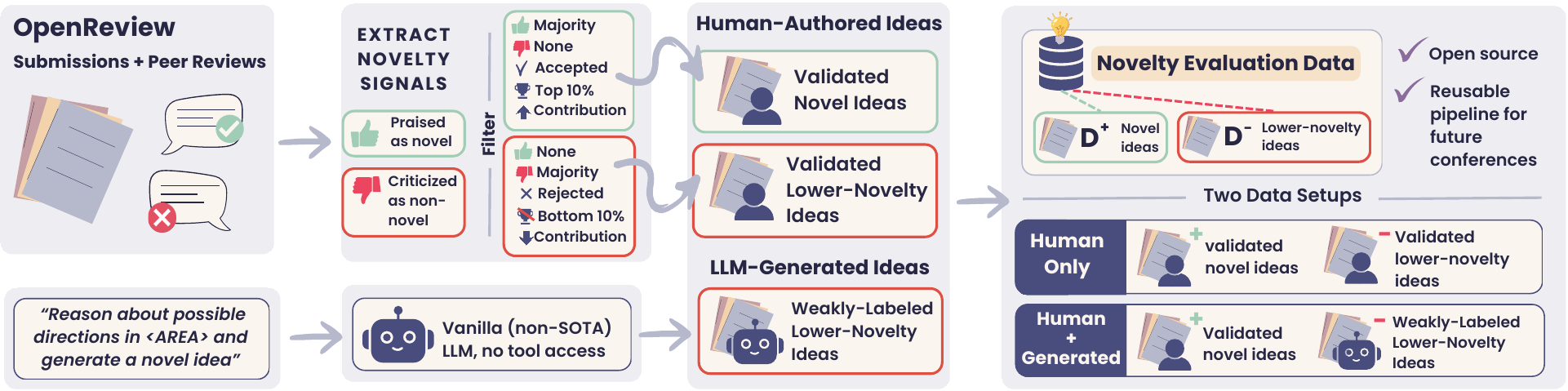}
    \caption{\textbf{Automatic data collection.} 
    We mine novelty labels from OpenReview, keeping only submissions whose reviewers \textit{agree} on their originality, and complement them with weakly labeled lower-novelty ideas from a non-SOTA LLM. The result is a novelty evaluation dataset with two setups, \humanonly and \humangen.
    }\label{fig:data-creation-overview}
\end{figure*}

To this end, we build a benchmark using an automated pipeline that mines high-precision novelty labels from peer reviews (Figure~\ref{fig:data-creation-overview}). Unlike prior work relying on coarse proxies like paper acceptance \cite{qiao2026innoevalresearchideaevaluation} or outdated novelty scores that venues no longer publish \cite{schopf2026ideanovelautomatedbenchmark}, we derive labels from the review 
itself. We isolate passages where reviewers explicitly affirm or dispute the originality of a contribution, retaining only submissions with unanimous consensus in the top or bottom decile of their venue. 
We complement these with LLM-generated ideas, sampled under settings very likely to yield lower novelty.

Using this benchmark, we find that novelty judges are strikingly \textit{brittle}. Changes to the evaluation setup, such as telling the judge that reviewers found one of two ideas novel, change a judge's verdict on more than half of the identical pairs it is shown and swing its accuracy by more than 50 points (Figure~\ref{fig:teaser}). Judges also behave differently on generated ideas than on human-authored ones, exposing a systematic gap in current evaluation practice: pairwise judges degrade once the comparison involves generated ideas, and grow markedly more brittle there. How brittle also depends on the judge, in ways that are hard to anticipate: in Figure~\ref{fig:teaser}, the steadier judge on human-authored ideas is the one that collapses on generated ones. The usual remedies, such as more reasoning or retrieval, help little and can even hurt.
We further show that costlier, more complex dedicated novelty evaluators do not necessarily give better verdicts.
Overall, LLM novelty judges are highly unstable, \textbf{calling into question the robustness of a large body of prior work 
that uses LLM judges 
to evaluate the novelty of 
generated scientific ideas}.

Our contributions are as follows:
\begin{itemize}[leftmargin=*, nosep]
    \item A 
    \textbf{\textit{controlled study of novelty judges}}, spanning six judge backbones, pointwise and pairwise judgments, and design choices 
    like
    prompt design, retrieval, reasoning effort, and idea format, on 
    human-written and LLM-generated ideas.
    \item An \textit{\textbf{automated, reusable pipeline}} that mines explicit novelty signals from peer reviews%
    , yielding two evaluation setups:
    \humanonly (strongest high-novelty vs. weakest low-novelty human papers) and \humangen (strongest high-novelty vs. ideas from a deliberately simple LLM generator). The pipeline can be rerun on new conference data, and we release all data and code.
    \item An empirical finding that the same judge 
    \textit{\textbf{flips its
    verdicts}} on the same instances 
    across
    plausible configurations. The effect can be 
    stronger
    on generated ideas, 
    where 
    judges are actually used, and common fixes (retrieval, more reasoning, costlier dedicated judges) offer limited help.

\end{itemize}

\section{Related Work}\label{sec:related_work}

\paragraph{Novelty evaluation of research ideas.}
Ideation systems span end-to-end agentic scientists \cite{Lu2026TowardsEA,Yamada2025TheAS}, multi-agent pipelines that debate and refine hypotheses \cite{Gottweis2025TowardsAA,su2025headsbetteroneimproved}, and literature-grounded ideators that search over or recombine prior work \cite{wang2024scimon,hu2024nova,li2024chainideasrevolutionizingresearch,baek2025researchagentiterativeresearchidea,radensky2026humanllmcompoundscientificideation}. Novelty is central to all of them, and its assessment is often delegated to LLM-based judges.

Large-scale evaluations of such novelty judges have so far used human-authored ideas. Examples are \citet{moussa2026scholarevalresearchideaevaluation,wu2026novbenchevaluatinglargelanguage}, who align judges with peer-review content, \citet{schopf2026ideanovelautomatedbenchmark,qiao2026innoevalresearchideaevaluation}, who extract novelty labels from review scores, and \citet{Lin2024EvaluatingAE}, who use a temporal proxy, treating the later of two arXiv papers published years apart as more novel. \citet{liu2026axiomaticbenchmarkevaluationscientific} forgo labels altogether, testing whether a novelty metric's score moves as expected when the pool of prior work is perturbed, e.g., by inserting the paper itself or removing the papers it cites. Work that does examine generated ideas rests on human annotation and is thus small and specific. \citet{Si2024CanLG} present a small dataset of mostly student-annotated ideas, focusing on a single generation system and $7$ LLM-related research topics. \citet{Shahid2025LiteratureGroundedNA} evaluate a literature-grounded novelty checker on a small set of ideas produced by one system \cite{radensky2026humanllmcompoundscientificideation}, \citet{sinhahajari2026limitsllmasjudgescientificnovelty} validate LLM judgments of generated research questions against expert annotation, and \citet{gupta2025allthatglitters} have annotators trace generated research documents back to prior work, targeting plagiarism rather than novelty. To our knowledge, no prior work systematically studies automatic novelty evaluation of generated ideas.

\paragraph{Sensitivity of LLM judges.} LLM judges are sensitive to design choices such as presentation order \cite{wang2023largelanguagemodelsfair}, pointwise versus pairwise formats \cite{liusie2024llm,tripathi2025pairwisepointwiseevaluatingfeedback}, and authority cues \cite{ye2025justice}. LLMs more broadly are swayed by 
external context \cite{xie2024adaptivechameleonstubbornsloth}.
 Such brittleness has seen little examination in novelty judgment, which we study here.

\section{Automated Benchmark Construction}

Our goal is to evaluate novelty judges on ideas spanning higher and lower levels of novelty, authored both by humans and by automated ideation systems. We collect ideas into two pools, such that ideas in one are expected to be more novel than those in the other (Figure~\ref{fig:data-creation-overview}). 
The described data collection pipeline scales to larger idea pools and can be rerun on future review cycles, keeping the benchmark evergreen. It is also not tied to ICLR: it applies to any venue that publishes free-text reviews, such as NeurIPS or ACL Rolling Review.

\subsection{Labeled Ideas Collection} \label{sub:idea_collection}

\paragraph{Validated novel ideas.} We extract novelty labels from ICLR 2026 submissions and their reviews. Unlike prior review-based datasets (\S\ref{sec:related_work}), which rely on acceptance decisions \cite{qiao2026innoevalresearchideaevaluation}, a coarse proxy that folds in other quality dimensions, or on novelty scores that venues no longer publish \cite{schopf2026ideanovelautomatedbenchmark}, we derive labels from the review text. Reviewers state verdicts on originality in prose, scattered across the strengths and weaknesses sections. We therefore prompt \texttt{claude-opus-4-6} to extract \textbf{novelty signals}: snippets where reviewers affirm or dispute the novelty of the \emph{contribution itself}.  We validate extraction on $50$ samples: an expert annotator agreed with the extracted polarity (positive/negative) $94$\% of the time, and no disagreement would flip a reviewer's polarity. Appendix~\ref{sub:human_data_collection} lists additional details on mining novelty labels from human reviews.

 We label a submission \textit{novel} when most reviewers praise its originality and none dispute it. It must also score at the top of the venue: accepted, in the top decile of its ICLR primary area by average rating, and above a fixed floor on average contribution score. We use its abstract as the idea.

\paragraph{Validated lower-novelty ideas.} The complementary pool inverts every criterion. Its reviewers must dispute the contribution's originality: a majority give negative novelty signals, and no reviewer gives a positive one. Its submission must sit at the bottom of the venue: rejected, with an average rating in the bottom decile of its ICLR primary area and an average contribution score below a fixed ceiling.

\paragraph{Weakly labeled lower-novelty ideas.} As discussed in \S\ref{sec:related_work}, novelty judges have been evaluated at scale only on \emph{human-authored} papers, while work on generated ideas, the target distribution when evaluating ideation systems, relies on small expert-annotated sets. Such annotations are slow and expensive to collect, and, once released, liable to leak into the training data of the very models later used as novelty judges. 
We therefore build the generated pool under an explicit weak-labeling assumption: \emph{every idea in the generated pool is less novel than every validated novel idea}. We design the pool to keep this assumption conservative. Its human side is the \emph{validated novel} pool described above: top-decile accepted ICLR papers whose reviewers, having read the full paper, unanimously affirmed the novelty of the contribution. Its generated side comes from \texttt{claude-sonnet-4-5}, a deliberately non-frontier model released over a year ago, which is given only the ICLR primary area of its human counterpart and asked to ``generate a novel idea'' in a single pass, with no literature access, tools, scaffold, or feedback. 
For the pairwise format, we consider each generated idea less novel than the top-decile human idea it is paired with, and for the pointwise format we treat generated ideas simply as not scientifically novel. 
We argue this is highly plausible given the state of technology in 2025 (see Appendix~\ref{sub:different_ideation_models} on newer generators).

As a check that the assumption is not obviously violated, a computer science professor with broad expertise across AI areas annotated $30$ \humangen pairs blind to source and with idea order shuffled, following the same instructions as our judge prompts. For each pair, the expert (1) chose the more novel idea and (2) decided whether the idea they ranked lower is novel, giving written justifications and citing prior work where relevant. We concentrated the sample where the assumption is most likely to fail. It included $25$ pairs our strongest pairwise judge (\texttt{claude-opus-4-6}) got wrong, and five controls it got right. The expert sided with the human idea in all $30$ pairs, with no ties, and judged none of the $30$ generated ideas novel, over roughly five hours of annotation. We present this as a sanity check on the hardest cases rather than a validation of every label (see Appendix~\ref{sub:expert_validation} for details and examples).

Finally, this labeling assumption has no bearing on our evaluation of \emph{instability}: two configurations that return different verdicts on the same pair cannot both be right, whatever its label.

\subsection{Benchmark Construction} \label{sub:task_description}
\paragraph{Pointwise vs. pairwise evaluation.} We consider the two common formats of novelty evaluation: (1) \textit{Pointwise} \cite{Shahid2025LiteratureGroundedNA,Yamada2025TheAS},  where the judge is given a single idea and decides whether it is novel enough to clear the bar of a top AI venue (binary classification), and (2) \textit{Pairwise} \cite{qiao2026innoevalresearchideaevaluation}, where the judge is given two ideas and decides which of the two is more novel (ranking).
The pointwise verdict is thus a thresholded judgment: an idea can be original in some respect and still be \texttt{not-novel} in this sense, as with a straightforward composition of known methods. Our judge prompt sets this bar by casting the judge as a senior reviewer at a venue such as ICLR and instructing it not to credit trivial innovation (Appendix~\ref{sub:prompted_judges}).
Beyond their prevalence, pointwise and pairwise evaluations are the atomic components underlying more complex protocols, such as majority voting and tournament-style ranking \cite{Si2024CanLG,Gottweis2025TowardsAA}.

\paragraph{Data setups.} We assemble the collected ideas into two setups, each a pair of pools: $\boldsymbol{\Dpos}$ (\textit{novel ideas}) and $\boldsymbol{\Dneg}$ (\textit{lower-novelty ideas}), populated so that an idea drawn from $\Dpos$ can be assumed more novel than one drawn from $\Dneg$. In \humanonly, $\Dpos$ holds the validated novel ideas and $\Dneg$ the validated lower-novelty ideas, so both sides are human-authored. In \humangen, $\Dpos$ is unchanged and $\Dneg$ holds the weakly labeled generated ideas. The two setups thus share $\Dpos$ and differ only in the source of $\Dneg$. Accordingly, $\Dneg$ does not denote a fixed set of ideas, but whichever pool fills the lower-novelty side of the setup at hand. Each data setup supports both pointwise and pairwise judgment. In \textit{pointwise}, every idea is one instance, labeled by its pool: $\Dpos$ ideas are \texttt{novel} and $\Dneg$ ideas \texttt{not-novel}, i.e., below the novelty bar of a top venue. In \textit{pairwise}, each idea in $\Dpos$ is matched with one from $\Dneg$ in the same ICLR primary area (e.g., \textit{generative models}), so the two sides are topically comparable. In \humanonly, where $\Dneg$ is the smaller pool, an idea may be matched more than once (Table~\ref{tab:benchmark_stats}, Appendix~\ref{sub:benchmark_setups}).

\begin{figure*}[!hbt]
    \centering
    \includegraphics[width=\textwidth]{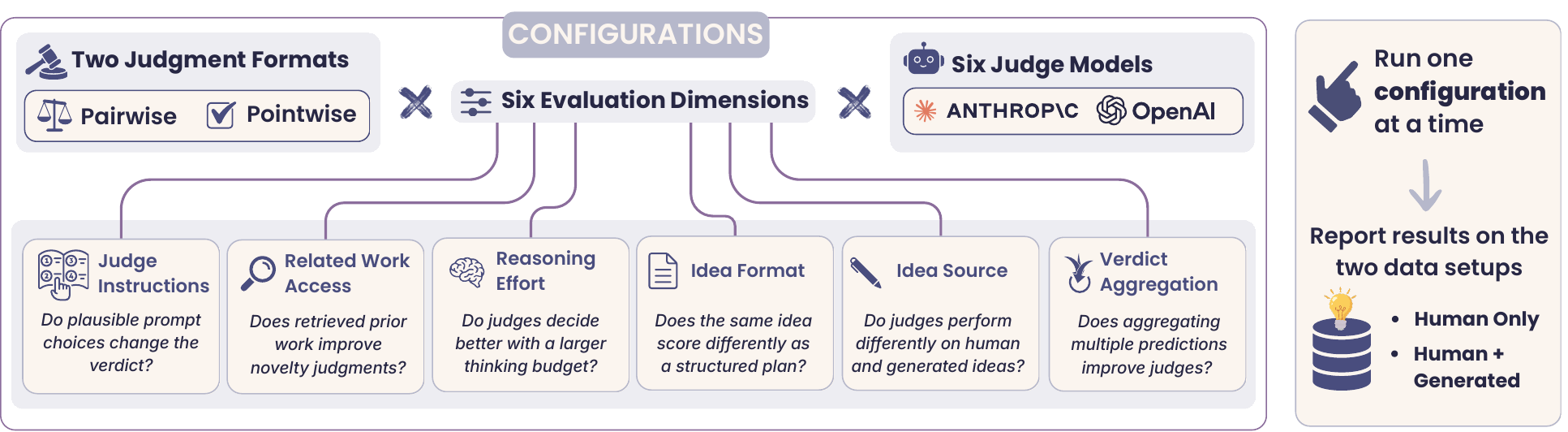}
    \caption{\textbf{Study overview.} We cross two judgment formats with six judge backbones and six evaluation dimensions. We run \textit{one configuration at a time} and report the results on both data setups, \humanonly and \humangen.}\label{fig:controlled-study-overview}
\end{figure*}

\paragraph{Idea format.} An idea in our benchmark is a single abstract, human-authored or generated. A submission's own abstract advertises experimental outcomes that an LLM ideator, with no execution, cannot produce without hallucinating results. We therefore apply \texttt{opus-4-6} to every human abstract to remove evaluation artifacts (e.g., numerical results, claims of superiority to specific baselines; see Prompt~\ref{prompt:remove-findings}, Appendix~\ref{sub:human_data_collection}). Generated ideas are emitted in this format directly.

\section{Experimental Setup} \label{sec:experimental_setup}

We perform a systematic study of novelty judges, spanning both judgment formats, a range of judge backbones, and several evaluation dimensions, on both data setups (Figure~\ref{fig:controlled-study-overview}). Each run modifies a \textit{single} design choice relative to \textit{\textbf{a reference configuration}}, holding the rest of the evaluation process fixed.\label{sub:reference_config} In the reference configuration, the judge receives a single idea (pointwise) or a pair of ideas (pairwise) and returns a novelty verdict with its reasoning, using \textit{high} reasoning effort (where exposed) and no literature search. To stabilize predictions, we aggregate multiple calls. Pointwise judges take the majority vote of three calls. Pairwise judges are called three times per presentation order ($A$ vs.\ $B$ and $B$ vs.\ $A$) to mitigate position bias \citep{wang2023largelanguagemodelsfair}, and the idea selected in more of the six calls wins. An even split is recorded as a \textbf{tie}, i.e., the judge cannot distinguish the two ideas. In \humangen, the reference $\Dneg$ is generated by \texttt{claude-sonnet-4-5} (\S\ref{sub:idea_collection}); since this generator fixes the lower-novelty side of the benchmark, we also treat it as a study parameter. Appendix~\ref{sub:study_details} summarizes the reference and its controlled changes (Table~\ref{tab:controlled_changes}), and Appendix~\ref{sub:prompted_judges} gives the judge prompts and implementation details.

\begin{figure*}[!htb]
    \centering
    \begin{subfigure}{\textwidth}
        \centering
        \includegraphics[width=\textwidth]{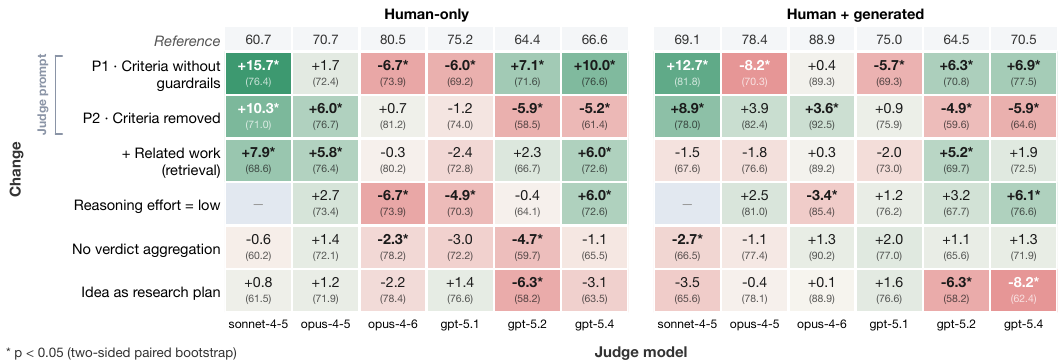}
        \caption{\textbf{Pointwise} judging, macro-F1.}\label{fig:ablation_pointwise}
    \end{subfigure}

    \vspace{1em}

    \begin{subfigure}{\textwidth}
        \centering
        \includegraphics[width=\textwidth]{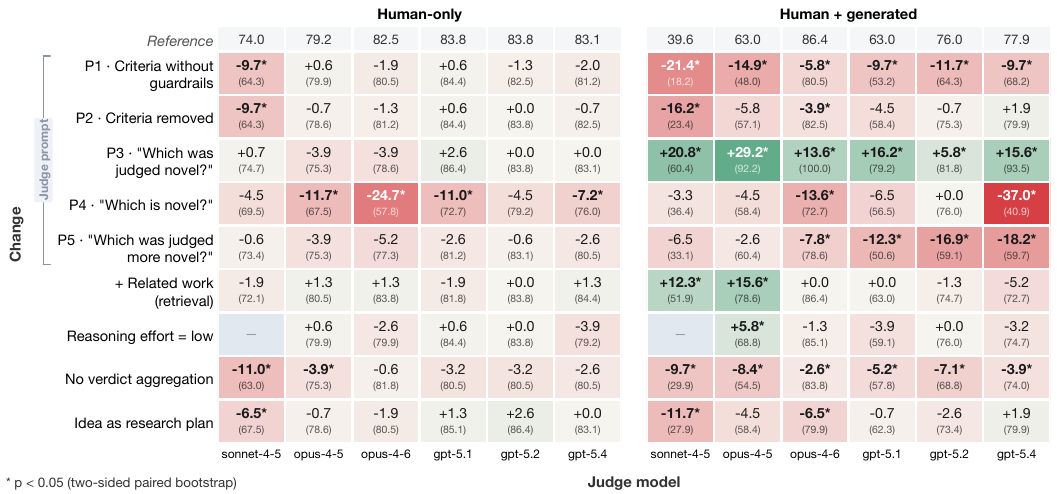}
        \caption{\textbf{Pairwise} judging, strict-accuracy (ties count as failures).}\label{fig:ablation_pairwise}
    \end{subfigure}
    \caption{\textbf{Judges are highly unstable} 
    under configuration changes,
    and models vary widely. Cells show the change from the reference (\S\ref{sub:reference_config}), with absolute values in parentheses (blank: unsupported).}
    \label{fig:ablation}
\end{figure*}

\subsection{Explored Evaluation Dimensions}\label{sub:eval_dimensions}
\paragraph{Judge prompt.} The instructions given to the judge. We explore six prompts: the reference, which casts the judge as a reviewer and gives explicit novelty criteria with guardrails against trivial contributions (Prompt~\ref{prompt:baseline-pairwise}), and five alternatives (P1--P5, Table~\ref{tab:controlled_changes}, Appendix~\ref{sub:study_details}). P1 and P2 weaken the novelty definition given in the reference prompt: P1 drops the warning against trivial or incremental contributions, and P2 removes the definition of novelty altogether. P3--P5 (a family of pairwise prompts inspired by \citet{Si2024CanLG}) ask the judge to choose between two ideas: P3 tells the judge that reviewers at a top AI conference found one idea novel and the other not, P4 omits this reviewer framing and asks which idea is novel, and P5 retains the reviewer framing but asks which idea was judged \emph{more} novel. 
The prompts are not equivalent, but each is a formulation a practitioner could reasonably choose for the same evaluation goal.

\paragraph{Retrieval.} Whether the judge sees related work alongside the idea. In the retrieval condition the judge additionally receives $5$ abstracts of related work, found by queries probing the novelty of the idea's main contributions and restricted to papers published before the submission deadline (Appendix~\ref{sub:study_implementation_details}). We also adjust the prompt, asking the judge to assess novelty against the retrieved work.

\paragraph{Reasoning effort.} The inference-time reasoning budget given to the judge. Assessing novelty requires comparing an idea against the judge's knowledge and deciding whether the overlap with prior work is substantial, which we expect to be reasoning-intensive. We test that expectation by lowering \texttt{reasoning\_effort} from \textit{high} to \textit{low} for the backbones that expose the parameter.

\paragraph{Idea format.} The surface form in which ideas are presented to the judge. Prior work presents ideas in different formats, from free-form abstracts \cite{wang2024scimon} to structured research plans \cite{Shahid2025LiteratureGroundedNA,Si2024CanLG}, typically without justifying the choice. We therefore test the effect of converting every idea into a two-field \textit{purpose} / \textit{mechanism} plan, which holds the content of the idea fixed and changes only its presentation.

\paragraph{Verdict aggregation.} The number of judge calls we pool into one decision. Aggregation is a simple but costly way to stabilize verdicts \cite{wang2023largelanguagemodelsfair}. We test whether dividing the aggregated calls by $3$ (calling pointwise judges once, pairwise judges once per presentation order) degrades them.

\paragraph{Judge backbone.} We run our controlled study with a range of judge backbone models (\texttt{gpt-\allowbreak 5.\{1,2,4\}}, \texttt{claude-\allowbreak sonnet-\allowbreak 4-5}, \texttt{claude-\allowbreak opus-\allowbreak 4-\{5,6\}}), all with training cutoffs predating the ICLR 2026 submission deadline so that the evaluated ideas fall outside their training data.
The novelty labels are further still out of reach: they are derived from reviews that were only released on OpenReview on 2025-11-12, roughly two months after the deadline. Thus, even if an idea reached a judge's training data through an earlier preprint, the reviewers' verdict on its novelty could not have.

\subsection{Evaluation Metrics} 
For pointwise evaluation we report \textbf{F1} per class (novel / not-novel) and a macro average. For pairwise, we report two accuracy variants that differ in tie handling: (1) \textbf{soft-accuracy}: ties count as $0.5$; (2) \textbf{strict-accuracy}: ties receive no credit. We mark with * changes that differ significantly from the reference configuration ($p < 0.05$, two-sided non-parametric paired bootstrap; Appendix~\ref{sub:study_implementation_details}).

\section{Results}\label{sec:results}

Figure~\ref{fig:ablation} reports the controlled-study results.
The first thing to note is that judges generally underperform, and rarely approach perfect results on this task. This happens even in the \humanonly setup, where $\Dpos$ and $\Dneg$ are drawn from opposite ends of the ICLR rating distribution and should be maximally separable.
Performance is also highly unstable under configuration changes. Removing the guardrails from the novelty criteria, or dropping the criteria entirely (P1--P2), generally degrades pairwise judges while helping some pointwise ones. On pairwise \humangen, replacing the reference prompt with one that asks which idea human reviewers judged novel (P3) improves every judge (by over 20 points in some cases). Yet deleting the sentence about reviewers (P4) makes \texttt{gpt-5.4} collapse below $50\%$. Because both prompts judge the same pairs,
the $52.6$-point gap means they disagree on at least $52.6$\% of pairs (55.8\% in practice), so this instability is independent of our labels.

To rule out the possibility that judges are reacting to stylistic differences between human-written and LLM-generated ideas rather than to their novelty, we rewrite all ideas into a shared plan format and rerun a subset of the changes that produced the largest shifts. The direction of the shifts is unchanged, and instability even grows: some changes flip the verdicts on more than two thirds of the examples, and P4 now costs \texttt{gpt-5.4} $24$ points even on \humanonly (Figure~\ref{fig:plan_format}, Appendix~\ref{sub_additional_ablation_results}).

These shifts also do not move models together. The same change can be neutral or beneficial for one judge but destructive for another. This is clearest in pointwise judging (e.g., P1--P2), but also holds pairwise: retrieval is a large gain for \texttt{claude-opus-4-5} on \humangen while neutral for others. A configuration validated on one judge thus carries no guarantee for another.

Design choices that prior work commonly reaches for, like increased reasoning \cite{Saha2025LearningTPA} or retrieval \cite{Shahid2025LiteratureGroundedNA,Lin2024EvaluatingAE}, have a limited and unexpected effect. Lowering the reasoning effort from \texttt{"high"} to \texttt{"low"} costs at most $7$ points, and is usually neutral. Surprisingly, some models even benefit from a smaller reasoning budget. Retrieval is similarly underwhelming: it yields a statistically significant gain of at most $16$ points (less than replacing the reference prompt with P3), only for some models, and never brings the absolute numbers near $100$. This remains true when $\Dneg$ is generated by other backbones, where retrieval even significantly degrades performance in some cases (Figure~\ref{fig:retrieval_by_generator}).

The effect of the same prompt change also depends on the judge and the data.
Switching from P3 to P4 already shifts strict-accuracy on \humanonly by $21$ points for \texttt{claude-opus-4-6}, but only about $7$ for \texttt{gpt-5.4}. On \humangen, the shift grows for both judges, and \texttt{gpt-5.4} becomes the more sensitive one ($52.6$ points). 
Which judge looks robust thus depends on the data, and validating on human-authored ideas can understate how much a prompt matters on generated ones.

Interestingly, removing instructions against trivial or additive novelty from the prompt (P1) harms every judge once AI-generated ideas are involved, while being mostly neutral on human-authored pairs. This might indicate that LLM-generated ideas lean more towards this style of ideation. 

Across ideation models generating $\Dneg$, judges generally degrade once the lower-novelty ideas are LLM-generated rather than human-authored, with performance below chance for some backbones (Figure~\ref{fig:negatives_source}, Appendix~\ref{sub:different_ideation_models}). Older generators are no easier to judge: ideas from \texttt{gpt-5} hurt judges about as much as those from \texttt{gpt-5.4}.

Another concern is the judges’ tie rate: how often a pairwise judge gives an uncertain verdict (\S\ref{sub:reference_config}). Depending on the configuration, even strong judges like \texttt{claude-\allowbreak opus-\allowbreak 4-6} can tie on $35\%$ of pairs, and weaker ones like \texttt{claude-\allowbreak sonnet-\allowbreak 4-5} on \textit{more than half}. Ignoring ties or counting them as half-correct can substantially change how evaluation systems are ranked (soft-accuracy results in Figure~\ref{fig:ablation_pairwise_soft}, Appendix~\ref{sub_additional_ablation_results}). Judges tie more often on \humangen than on \humanonly, roughly twice as often for \texttt{claude-\allowbreak opus-\allowbreak 4-5} and \texttt{claude-\allowbreak sonnet-\allowbreak 4-5}, indicating higher uncertainty there. Turning off verdict aggregation adds further uncertain predictions, more than doubling tie rates in certain cases (Table~\ref{tab:tie_rates}, Appendix~\ref{sub_additional_ablation_results}). Together with the tendency to tie more on generated data, this explains the drop it causes on \humangen.

We also observe that, in line with prior work \cite{raina2024llm,liusie2024llm}, pairwise judging is more stable than pointwise judging, but only on \humanonly. On the same data, pointwise judges are as unstable as on \humangen.

\paragraph{Comparing ideation systems.} A common use of a novelty judge is to compare two ideation systems, a setting with no gold label. To test it, we pit the ideas of a stronger backbone against those of a weaker one from the same family (Appendix~\ref{sub:ai_vs_ai}), and count the number of verdicts changed. Changing the prompt alone flips a quarter to a third of the verdicts on average, and moves the share of pairs awarded to the stronger system by up to $29$ points.

\subsection{Retrieval's Limitations} \label{subsec:retrieval_limitations}

\paragraph{Preliminary qualitative analysis.} Surprisingly, retrieval yields limited gains, if any (Figure~\ref{fig:ablation}). To understand why, we inspect $10$ cases where retrieval flips a correct pointwise verdict to a wrong one, keeping only flips where all three aggregated calls agree on each side, ruling out noise (Appendix~\ref{sub:retrieval_qualitative}). We find that retrieved papers can lead the judge to dismiss contributions it had considered novel, even when nothing contradicts their novelty. In one case, the judge without retrieval deemed an idea combining two concepts novel, agreeing with the reviewers. Given papers presenting each concept separately, it dismissed the same idea as ``a direct combination of existing ingredients\ldots''. Retrieval can also mislead the other way, overriding the judge's knowledge. Without retrieval, the judge correctly described an idea from $\Dneg$ as ``a broad composition of known components'', but with retrieval it concluded the idea ``\ldots does appear novel in its overall problem formulation and synthesis \textit{relative to the provided related work}\ldots''. This matters, as many pipelines ask novelty judges to assess ideas against retrieved work \cite{Shahid2025LiteratureGroundedNA,moussa2026scholarevalresearchideaevaluation}.

\paragraph{Stronger retrieval.} A possible explanation is that our retrieval pipeline is too weak, and a stronger retriever would let judges ground their verdicts in the right prior work. We test this by exploring a much stronger retrieval-augmented judge.
We instantiate a \emph{strong agentic RAG judge} on top of \texttt{gpt-5.6-sol}, a newer and stronger backbone than any judge so far, run at \texttt{reasoning\_effort=xhigh} and equipped with a web-search tool. Unlike the standard pipeline, which hands judges five pre-retrieved abstracts, this judge searches the literature iteratively, with no cap on searches or inspected results. 
As in our standard pipeline, search is restricted to papers published before the submission deadline (Appendix~\ref{sub:strong_retrieval_judge}).
\begin{figure}[t]
    \centering
    \includegraphics[width=\columnwidth]{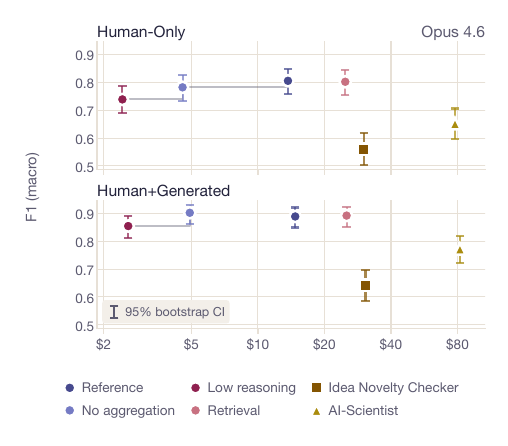}
    \caption{
        \textbf{Cost vs.\ pointwise macro-F1} for \texttt{opus-4-6}, comparing prompted-judge configurations against the dedicated judges. The Pareto front (line) contains \emph{only} prompted judges: the \emph{cheapest} prompted configuration outscores dedicated judges at up to $32\times$ lower cost.
        }
    \label{fig:pareto_pointwise_opus}
\end{figure}
We compare it on \humangen to judges using the standard pipeline, all called once per idea for parity (Figure~\ref{fig:retrieval_faceoff} in the appendix). The agentic judge reaches $0.78$ macro-F1, $11$ points behind \texttt{claude-opus-4-6} and within noise of \texttt{claude-opus-4-5} and \texttt{gpt-5.4}, at $10\times$ the cost. The modest retrieval gains in Figure~\ref{fig:ablation} are thus not an artifact of our pipeline.

\subsection{Dedicated Novelty Judges}\label{sub:existing_novelty_eval}

Our study focuses on prompted novelty judges, off-the-shelf LLMs instructed to return a novelty verdict, since this is how most ideation systems assess novelty \cite{li2024chainideasrevolutionizingresearch,hu2024nova,shen2026unlockingllmcreativityscience}. Prior work also proposes dedicated novelty judges, which spend far more compute per decision. We test whether this pays off for two of them: \textbf{Idea Novelty Checker} \cite{Shahid2025LiteratureGroundedNA}, a literature-grounded pipeline that retrieves, filters, and re-ranks candidate papers before judging the idea against them, and \textbf{AI-Scientist} \cite{Yamada2025TheAS}, a ReAct-style agent that iteratively searches Semantic Scholar until it reaches a verdict. We run each with \texttt{claude-opus-4-6} and \texttt{gpt-5.4} under its original settings (one call per decision, without verdict aggregation). Since both are pointwise, we compare them to prompted judges in the pointwise format only, measuring cost in USD over the full evaluation set (details in Appendix~\ref{sec:judge_implementation_dits}, per-evaluator results in Appendix~\ref{sub:existing_evaluators}).

Figure~\ref{fig:pareto_pointwise_opus} plots the cost-quality tradeoff for \texttt{claude-opus-4-6} (see Figure~\ref{fig:pareto_pointwise_gpt} for \texttt{gpt-5.4}). The extra machinery does not pay off: in every setup and backbone, our \emph{cheapest} prompted configuration outscores both dedicated judges by $8$ to $29$ macro-F1 points, while costing up to $32\times$ less.

\section*{Conclusions}

We ran a controlled study of LLM novelty judges, using an automatic pipeline that mines novelty labels from 
peer-review text
and pairs human-authored with LLM-generated ideas. Judges are brittle: plausible prompt edits swing pairwise accuracy by over $50$ points, and the same edit that helps one backbone sinks another below chance. Even on human-authored ideas, where judges are usually validated, a single edit moves a strong judge by over $20$ points, and which judge looks most robust can flip once generated ideas are involved.
 The usual remedies (retrieval, more reasoning, costlier dedicated judges) help little.
Pairwise judges are least stable in the setting they are built for: large shifts are far more common on \humangen than on \humanonly, and judges grow less certain on generated data.
Validating a judge on human-authored papers, or against a single ideation system, says little about how it will behave on the generated ideas it is asked to score. Until evaluation protocols account for these sensitivities, automated novelty assessments should be interpreted with 
caution.

\section*{Limitations}

\paragraph{Weak labels for generated ideas.} The lower-novelty side of \humangen carries no observed novelty label. Its labels rest instead on a weak-labeling assumption: a top-decile ICLR submission whose reviewers agree is original is more novel than an idea produced in a single scaffold-free, tool-free pass of a deliberately non-frontier model. We argue for this ordering in \S\ref{sub:idea_collection}, but cannot verify it for the entire dataset, and a generated idea that occasionally clears the bar would add label noise. Such noise could blur which configuration is more accurate, but not our instability findings. If two configurations score differently on the same examples, they disagree on some of them, and at most one can be right on each, whatever the true label. To check that the assumption is not obviously violated, we ran a blind expert study. Because our pairs span many ICLR areas, judging them requires broad expertise in AI, so we recruited a single computer science professor with that breadth. They annotated 30 pairs in roughly five hours, most of them pairs our strongest judge got wrong, and agreed with our labels on every one (Appendix~\ref{sub:expert_validation}). This makes a systematic violation of the assumption improbable. The assumption is strongest for the non-frontier generator used in most of our experiments (\texttt{claude-sonnet-4-5}). Stronger generators are more likely to produce ideas that clear the bar, so results that regenerate the lower-novelty pool with them warrant more caution.

\paragraph{Partial coverage of the design space.} Our controlled study varies one design dimension at a time relative to a fixed reference configuration. We do probe interactions between design choices, but only in a few places: the retrieval experiment across ideation backbones, for instance, uses a single judge per model family (Appendix~\ref{sub_additional_ablation_results}, Figure~\ref{fig:retrieval_by_generator}).  The full cross-product of judges, ideation backbones, and design choices grows quickly, so we intentionally picked only a few cross-interaction experiments in order to keep costs manageable. For reference, the experiments described here cost roughly \$15{,}000 in API spend at $2026$ list prices, including preliminary and discarded runs.

\paragraph{Other evaluation criteria.} We focus on novelty because it is a widely used and thoroughly studied evaluation criterion. However, novelty is only one of several important criteria for evaluating ideas. Others include estimated impact \cite{li2026mirai} and surprisal \cite{agarwal2026autodiscoveryopenendedscientificdiscovery}. We believe many of these attributes are intertwined, and future work should study them as well.

\section*{Ethical Considerations}
Our benchmark is built from ICLR $2026$ submissions and reviews obtained through the public OpenReview API, and used in accordance with its terms. Both annotators in our validation studies (\S\ref{subsub:novelty-sig-validation}, \S\ref{sub:expert_validation}) are volunteers. Part of our data consists of LLM-generated ideas, which we release clearly marked as synthetic and unvalidated. Given that our findings show current novelty judges to be brittle, we caution against deploying them in consequential settings, and in particular as a substitute for human peer review. To support transparency and reproducibility, we release our data, prompts, and code under an open license. We used LLM-based tools for parts of our data collection pipeline (\S\ref{sub:human_data_collection}) and for minor writing and coding assistance.

\bibliography{custom}

@Article{Si2024CanLG,
 author = {Chenglei Si and Diyi Yang and Tatsunori Hashimoto},
 booktitle = {arXiv.org},
 journal = {ArXiv},
 title = {Can LLMs Generate Novel Research Ideas? A Large-Scale Human Study with 100+ NLP Researchers},
 volume = {abs/2409.04109},
 year = {2024}
}

@article{Shahid2025LiteratureGroundedNA,
  title={Literature-Grounded Novelty Assessment of Scientific Ideas},
  author={Simra Shahid and Marissa Radensky and Raymond Fok and Pao Siangliulue and Daniel S. Weld and Tom Hope},
  journal={ArXiv},
  year={2025},
  volume={abs/2506.22026},
  url={https://api.semanticscholar.org/CorpusID:280012372}
}

@misc{shen2026unlockingllmcreativityscience,
      title={Unlocking LLM Creativity in Science through Analogical Reasoning}, 
      author={Andrew Shen and Shaul Druckmann and James Zou},
      year={2026},
      eprint={2605.11258},
      archivePrefix={arXiv},
      primaryClass={cs.AI},
      url={https://arxiv.org/abs/2605.11258}, 
}

@article{Lu2026TowardsEA,
  title={Towards end-to-end automation of AI research},
  author={Chris Lu and Cong Lu and Robert Tjarko Lange and Yutaro Yamada and Shengran Hu and Jakob Foerster and David Ha and Jeff Clune},
  journal={Nature},
  year={2026},
  volume={651},
  pages={914 - 919},
  url={https://api.semanticscholar.org/CorpusID:286823959}
}

@article{Yamada2025TheAS,
  title={The AI Scientist-v2: Workshop-Level Automated Scientific Discovery via Agentic Tree Search},
  author={Yutaro Yamada and Robert Tjarko Lange and Cong Lu and Shengran Hu and Chris Lu and Jakob Foerster and Jeff Clune and David Ha},
  journal={ArXiv},
  year={2025},
  volume={abs/2504.08066},
  url={https://api.semanticscholar.org/CorpusID:277741107}
}

@article{Gottweis2025TowardsAA,
  title={Towards an AI co-scientist},
  author={Juraj Gottweis and Wei-Hung Weng and Alexander Daryin and Tao Tu and Anil Palepu and Petar Sirkovic and Artiom Myaskovsky and Felix Weissenberger and Keran Rong and Ryutaro Tanno and Khaled Saab and Dan George Popovici and Jacob Blum and Fan Zhang and Katherine Chou and Avinatan Hassidim and Burak Gokturk and Amin Vahdat and Pushmeet Kohli and Yossi Matias and Andrew Carroll and Kavita Kulkarni and Nenad Toma{\v s}ev and Yuan Guan and Vikram Dhillon and Eeshit Dhaval Vaishnav and Byron Lee and Tiago R D Costa and Jos'e R Penad'es and Gary Peltz and Yunhan Xu and Annalisa Pawlosky and Alan Karthikesalingam and Vivek Natarajan},
  journal={ArXiv},
  year={2025},
  volume={abs/2502.18864},
  url={https://api.semanticscholar.org/CorpusID:276617649}
}

@misc{schopf2026ideanovelautomatedbenchmark,
      title={Is this Idea Novel? An Automated Benchmark for Judgment of Research Ideas}, 
      author={Tim Schopf and Michael Färber},
      year={2026},
      eprint={2603.10303},
      archivePrefix={arXiv},
      primaryClass={cs.CL},
      url={https://arxiv.org/abs/2603.10303}, 
}

@misc{moussa2026scholarevalresearchideaevaluation,
      title={ScholarEval: Research Idea Evaluation Grounded in Literature}, 
      author={Hanane Nour Moussa and Patrick Queiroz Da Silva and Daniel Adu-Ampratwum and Alyson East and Zitong Lu and Nikki Puccetti and Mingyi Xue and Huan Sun and Bodhisattwa Prasad Majumder and Sachin Kumar},
      year={2026},
      eprint={2510.16234},
      archivePrefix={arXiv},
      primaryClass={cs.AI},
      url={https://arxiv.org/abs/2510.16234}, 
}

@misc{qiao2026innoevalresearchideaevaluation,
      title={InnoEval: On Research Idea Evaluation as a Knowledge-Grounded, Multi-Perspective Reasoning Problem}, 
      author={Shuofei Qiao and Yunxiang Wei and Xuehai Wang and Bin Wu and Boyang Xue and Ningyu Zhang and Hossein A. Rahmani and Yanshan Wang and Qiang Zhang and Keyan Ding and Jeff Z. Pan and Huajun Chen and Emine Yilmaz},
      year={2026},
      eprint={2602.14367},
      archivePrefix={arXiv},
      primaryClass={cs.CL},
      url={https://arxiv.org/abs/2602.14367}, 
}

@misc{gupta2025allthatglitters,
      title={All That Glitters is Not Novel: Plagiarism in AI Generated Research}, 
      author={Tarun Gupta and Danish Pruthi},
      year={2025},
      eprint={2502.16487},
      archivePrefix={arXiv},
      primaryClass={cs.CL},
      url={https://arxiv.org/abs/2502.16487}, 
}

@Article{Lin2024EvaluatingAE,
 author = {Ethan Lin and Zhiyuan Peng and Yi Fang},
 booktitle = {arXiv.org},
 journal = {ArXiv},
 title = {Evaluating and Enhancing Large Language Models for Novelty Assessment in Scholarly Publications},
 volume = {abs/2409.16605},
 year = {2024}
}

@article{hu2024nova,
  title={Nova: An iterative planning and search approach to enhance novelty and diversity of llm generated ideas},
  author={Hu, Xiang and Fu, Hongyu and Wang, Jinge and Wang, Yifeng and Li, Zhikun and Xu, Renjun and Lu, Yu and Jin, Yaochu and Pan, Lili and Lan, Zhenzhong},
  journal={arXiv preprint arXiv:2410.14255},
  year={2024}
}

@misc{baek2025researchagentiterativeresearchidea,
      title={ResearchAgent: Iterative Research Idea Generation over Scientific Literature with Large Language Models}, 
      author={Jinheon Baek and Sujay Kumar Jauhar and Silviu Cucerzan and Sung Ju Hwang},
      year={2025},
      eprint={2404.07738},
      archivePrefix={arXiv},
      primaryClass={cs.CL},
      url={https://arxiv.org/abs/2404.07738}, 
}

@misc{li2024chainideasrevolutionizingresearch,
      title={Chain of Ideas: Revolutionizing Research Via Novel Idea Development with LLM Agents}, 
      author={Long Li and Weiwen Xu and Jiayan Guo and Ruochen Zhao and Xingxuan Li and Yuqian Yuan and Boqiang Zhang and Yuming Jiang and Yifei Xin and Ronghao Dang and Deli Zhao and Yu Rong and Tian Feng and Lidong Bing},
      year={2024},
      eprint={2410.13185},
      archivePrefix={arXiv},
      primaryClass={cs.AI},
      url={https://arxiv.org/abs/2410.13185}, 
}

@misc{radensky2026humanllmcompoundscientificideation,
      title={Human-LLM Compound System for Scientific Ideation through Facet Recombination and Novelty Evaluation}, 
      author={Marissa Radensky and Simra Shahid and Raymond Fok and Pao Siangliulue and Tom Hope and Daniel S. Weld},
      year={2026},
      eprint={2409.14634},
      archivePrefix={arXiv},
      primaryClass={cs.HC},
      url={https://arxiv.org/abs/2409.14634}, 
}

@misc{wang2023largelanguagemodelsfair,
      title={Large Language Models are not Fair Evaluators}, 
      author={Peiyi Wang and Lei Li and Liang Chen and Zefan Cai and Dawei Zhu and Binghuai Lin and Yunbo Cao and Qi Liu and Tianyu Liu and Zhifang Sui},
      year={2023},
      eprint={2305.17926},
      archivePrefix={arXiv},
      primaryClass={cs.CL},
      url={https://arxiv.org/abs/2305.17926}, 
}

@misc{chen2026measuringgaphumanllm,
      title={Measuring the Gap Between Human and LLM Research Ideas}, 
      author={Ziyu Chen and Yilun Zhao and Arman Cohan},
      year={2026},
      eprint={2607.01233},
      archivePrefix={arXiv},
      primaryClass={cs.CL},
      url={https://arxiv.org/abs/2607.01233}, 
}

@inproceedings{raina2024llm,
  title={Is llm-as-a-judge robust? investigating universal adversarial attacks on zero-shot llm assessment},
  author={Raina, Vyas and Liusie, Adian and Gales, Mark},
  booktitle={Proceedings of the 2024 Conference on Empirical Methods in Natural Language Processing},
  pages={7499--7517},
  year={2024}
}

@inproceedings{liusie2024llm,
  title={LLM comparative assessment: Zero-shot NLG evaluation through pairwise comparisons using large language models},
  author={Liusie, Adian and Manakul, Potsawee and Gales, Mark},
  booktitle={Proceedings of the 18th Conference of the European Chapter of the Association for Computational Linguistics (Volume 1: Long Papers)},
  pages={139--151},
  year={2024}
}

@article{Saha2025LearningTPA,
  title={Learning to Plan \& Reason for Evaluation with Thinking-LLM-as-a-Judge},
  author={Swarnadeep Saha and Xian Li and Marjan Ghazvininejad and Jason Weston and Tianlu Wang},
  journal={ArXiv},
  year={2025},
  volume={abs/2501.18099},
  url={https://arxiv.org/pdf/2501.18099.pdf}
}

@misc{su2025headsbetteroneimproved,
      title={Many Heads Are Better Than One: Improved Scientific Idea Generation by A LLM-Based Multi-Agent System}, 
      author={Haoyang Su and Renqi Chen and Shixiang Tang and Zhenfei Yin and Xinzhe Zheng and Jinzhe Li and Biqing Qi and Qi Wu and Hui Li and Wanli Ouyang and Philip Torr and Bowen Zhou and Nanqing Dong},
      year={2025},
      eprint={2410.09403},
      archivePrefix={arXiv},
      primaryClass={cs.AI},
      url={https://arxiv.org/abs/2410.09403}, 
}

@inproceedings{wang2024scimon,
  title={Scimon: Scientific inspiration machines optimized for novelty},
  author={Wang, Qingyun and Downey, Doug and Ji, Heng and Hope, Tom},
  booktitle={Proceedings of the 62nd annual meeting of the association for computational linguistics (volume 1: long papers)},
  pages={279--299},
  year={2024}
}

@misc{wu2026novbenchevaluatinglargelanguage,
      title={NovBench: Evaluating Large Language Models on Academic Paper Novelty Assessment}, 
      author={Wenqing Wu and Yi Zhao and Yuzhuo Wang and Siyou Li and Juexi Shao and Yunfei Long and Chengzhi Zhang},
      year={2026},
      eprint={2604.11543},
      archivePrefix={arXiv},
      primaryClass={cs.CL},
      url={https://arxiv.org/abs/2604.11543}, 
}

@misc{liu2026axiomaticbenchmarkevaluationscientific,
      title={An Axiomatic Benchmark for Evaluation of Scientific Novelty Metrics}, 
      author={Miri Liu and ChengXiang Zhai},
      year={2026},
      eprint={2604.15145},
      archivePrefix={arXiv},
      primaryClass={cs.AI},
      url={https://arxiv.org/abs/2604.15145}, 
}

@misc{agarwal2026autodiscoveryopenendedscientificdiscovery,
      title={AutoDiscovery: Open-ended Scientific Discovery via Bayesian Surprise}, 
      author={Dhruv Agarwal and Bodhisattwa Prasad Majumder and Reece Adamson and Megha Chakravorty and Satvika Reddy Gavireddy and Aditya Parashar and Harshit Surana and Bhavana Dalvi Mishra and Andrew McCallum and Ashish Sabharwal and Peter Clark},
      year={2026},
      eprint={2507.00310},
      archivePrefix={arXiv},
      primaryClass={cs.LG},
      url={https://arxiv.org/abs/2507.00310}, 
}

@article{li2026mirai,
  title={MIRAI: Prediction and Generation of High-Impact Academic Research},
  author={Li, Alex and Jacobson, Joseph},
  journal={arXiv preprint arXiv:2606.05443},
  year={2026}
}

@misc{sinhahajari2026limitsllmasjudgescientificnovelty,
      title={On the Limits of LLM-as-Judge for Scientific Novelty Assessment}, 
      author={Soumitra Sinhahajari and Navonil Majumder and Soujanya Poria},
      year={2026},
      eprint={2606.12071},
      archivePrefix={arXiv},
      primaryClass={cs.DL},
      url={https://arxiv.org/abs/2606.12071}, 
}

@misc{tripathi2025pairwisepointwiseevaluatingfeedback,
      title={Pairwise or Pointwise? Evaluating Feedback Protocols for Bias in LLM-Based Evaluation}, 
      author={Tuhina Tripathi and Manya Wadhwa and Greg Durrett and Scott Niekum},
      year={2025},
      eprint={2504.14716},
      archivePrefix={arXiv},
      primaryClass={cs.LG},
      url={https://arxiv.org/abs/2504.14716}, 
}

@inproceedings{ye2025justice,
  title={Justice or prejudice? quantifying biases in llm-as-a-judge},
  author={Ye, Jiayi and Wang, Yanbo and Huang, Yue and Chen, Dongping and Zhang, Qihui and Moniz, Nuno and Gao, Tian and Geyer, Werner and Huang, Chao and Chen, Pin-Yu and others},
  booktitle={International Conference on Learning Representations},
  volume={2025},
  pages={102351--102390},
  year={2025}
}

@misc{xie2024adaptivechameleonstubbornsloth,
      title={Adaptive Chameleon or Stubborn Sloth: Revealing the Behavior of Large Language Models in Knowledge Conflicts}, 
      author={Jian Xie and Kai Zhang and Jiangjie Chen and Renze Lou and Yu Su},
      year={2024},
      eprint={2305.13300},
      archivePrefix={arXiv},
      primaryClass={cs.CL},
      url={https://arxiv.org/abs/2305.13300}, 
}

\appendix
\addtocontents{toc}{\protect\setcounter{tocdepth}{2}}
\addtocontents{lof}{\protect\setcounter{tocdepth}{2}}
\addtocontents{lot}{\protect\setcounter{tocdepth}{2}}
\setcounter{tocdepth}{2}

\clearpage
\renewcommand{\contentsname}{Appendix Contents}
\tableofcontents

\vspace{1em}
\listofPrompts

\vspace{1em}
\listoffigures

\vspace{1em}
\listoftables
\clearpage

\section{Additional Benchmark Details}

\subsection{Human Data Collection} \label{sub:human_data_collection}
\begin{promptfigure*}{Novelty signals extraction prompt.}{prompt:novelty-signals}
  \promptshortcaption{Novelty signals extraction}
  \renewcommand{\promptfontsize}{\promptfonttiny}%
  \promptfile{Novelty Signals Extraction Prompt}{prompts/extract_novelty_signals.jinja2}
\end{promptfigure*}

\begin{table*}[t]
\centering
\footnotesize
\setlength{\tabcolsep}{5pt}
\renewcommand{\arraystretch}{1.2}
\newcommand{\sig}[1]{\textbf{#1}}
\begin{tabular}{@{}p{0.27\textwidth}p{0.69\textwidth}@{}}
\toprule
\textbf{ICLR 2026 submission} & \textbf{Extracted reviewer novelty signal examples} \\
\midrule
\multicolumn{2}{@{}l}{\textcolor{palDeepTeal}{\textbf{Positive novelty signals}}} \\
\addlinespace[2pt]
\textit{Image Quality Assessment for Embodied AI} &
``The paper identifies and clearly defines a \sig{completely new, critical, and timely research problem}'' \\
\addlinespace[4pt]
\textit{Quotient-Space Diffusion Models} &
``It is the \sig{first work I have seen} that attempts to \sig{constrain the diffusion process within the quotient space}'' \\
\addlinespace[4pt]
\textit{Critical Confabulation: Can LLMs Hallucinate for Social Good?} &
``this paper is \sig{so unlike most of the ICLR paper I've seen in recent years}, especially with its \sig{use of historical archival data}\ldots'' \\
\midrule
\multicolumn{2}{@{}l}{\textcolor{palTerracotta}{\textbf{Negative novelty signals}}} \\
\addlinespace[2pt]
\textit{DFCA: Decentralized Federated Clustering Algorithm} &
``While the theoretical contribution is \sig{not highly novel}, the paper provides a useful convergence discussion\ldots'' \\
\addlinespace[4pt]
\textit{NoLoRA: Nonlinear Low-Rank Adaptation for Parameter-Efficient Fine-Tuning} &
``The proposal is \sig{extremely close to NEAT} (Zhong et al., 2025). Both methods replace LoRA's linear update with a nonlinear mapping\ldots'' \\
\addlinespace[4pt]
\textit{Consistent 3D Object Detection with Active LLM Reasoning} &
``The contribution is \sig{primarily procedural integration of existing components} (YOLOE + CLIP + GPT-4V + SAM2), with \sig{no novel learning formulation or analytical insight}\ldots'' \\
\bottomrule
\end{tabular}
\caption[Novelty signals in ICLR reviews]{\textbf{Novelty signals extracted from ICLR 2026 reviews.}}
\label{tab:novelty_signals}
\end{table*}

We begin human-authored idea collection by fetching all ICLR $2026$ submissions and their corresponding reviews. We then apply \texttt{claude-opus-4-6} to extract novelty signals from the raw weaknesses/strengths review sections. Prompt~\ref{prompt:novelty-signals} presents the prompt used for novelty signal extraction, and Table~\ref{tab:novelty_signals} shows extraction examples. Next, we apply filters and divide the submissions into two groups as follows:
\begin{itemize}
    \item \textbf{High-novelty papers}: \textit{Accepted} submissions that have received an average rating within the \textit{top} 10\% of their ICLR area (and that is $\geq 6$), and an average contribution score $\geq 3$. Additionally, these submissions must have \textit{no negative novelty signals} and must receive \textit{positive} novelty signals from the \textit{majority} of reviewers.
    \item \textbf{Low-novelty papers}: \textit{Rejected} submissions that have received an average rating within the \textit{bottom} 10\% of their ICLR area (and that is $\leq 3.5$), and an average contribution score $\leq 2$. Additionally, these submissions must have \textit{no positive novelty signals} and must receive \textit{negative} novelty signals from the \textit{majority} of reviewers.
\end{itemize}

We additionally exclude submissions whose primary ICLR area is \textit{other topics in machine learning} (i.e., none of the listed areas). This category aggregates a heterogeneous mix of unrelated subfields, which makes novelty comparisons between its submissions noisy.

\subsubsection{Novelty Signal Extraction Validation}\label{subsub:novelty-sig-validation}
We validate the extracted novelty signals with a manual annotation study. A PhD student with NLP expertise reviewed signals extracted from $50$ different ideas in \humanonly ($25$ validated novel, $25$ validated lower-novelty).

\paragraph{Sampling and setup.} We randomly sampled 25 ideas from each pool and one extracted signal per idea, so that no idea contributes more than one signal. Items were shuffled and presented blind: the annotator saw the idea's ICLR primary area, the idea text (the abstract given to the judges), and the signal, but not the extracted polarity, the paper's decision, or its review scores. For each item, the annotator answered two questions:
\begin{enumerate}
    \item \textbf{Polarity.} Does the snippet state that the contribution is novel (\textit{positive}), not novel (\textit{negative}), or does it judge something other than novelty (\textit{not about novelty})?
    \item \textbf{Missing from the abstract.} Is what the reviewer calls novel (or not novel) missing from the abstract?
\end{enumerate}

\paragraph{Results.} The annotator's polarity matched the extracted polarity for 47 of the 50 signals (94\% precision: 96\% for positive and 92\% for negative signals), with Cohen's $\kappa = 0.89$ (almost perfect agreement) when treating \textit{not about novelty} as a third class. Two of the three disagreements were snippets judged not to be about novelty, and the third was a negative snippet the annotator read as positive. None of the three would flip the overall polarity of the reviewer (Table~\ref{tab:signal_extraction_errors}). Finally, only 1 of the 47 correctly extracted signals (2.1\%) judged an aspect of the contribution that is missing from the abstract, suggesting that the reviewers' novelty verdicts are largely recoverable from the ideas the judges see.

\begin{table*}[t]
\centering
\footnotesize
\setlength{\tabcolsep}{5pt}
\renewcommand{\arraystretch}{1.2}
\newcommand{\sig}[1]{\textbf{#1}}
\newcommand{\note}[1]{\textcolor{palDarkCharcoal!70}{\textit{#1}}}
\begin{tabularx}{\textwidth}{@{}Y{1} Y{1}@{}}
\toprule
\textbf{Disputed signal} & \textbf{Other signals (same reviewer)} \\
\midrule
\rowcolor{palWarmTan!18}[0pt][0pt]
\multicolumn{2}{@{}l@{}}{\textbf{ICLR 2026 submission:} \textit{Gym4ReaL}\qquad \textbf{Extracted:} \textcolor{palTerracotta}{Negative}\quad \textbf{Annotator:} Not about novelty} \\
``The paper does not make a compelling case for how these environments \sig{fill known gaps in benchmark coverage} or offer unique real-world relevance.'' &
``The overall contribution is therefore \sig{not well differentiated from prior work}.'' \note{(still conveys the extracted negative polarity)} \\
\addlinespace[3pt]
\rowcolor{palWarmTan!18}[0pt][0pt]
\multicolumn{2}{@{}l@{}}{\textbf{ICLR 2026 submission:} \textit{TD-MoE}\qquad \textbf{Extracted:} \textcolor{palDeepTeal}{Positive}\quad \textbf{Annotator:} Not about novelty} \\
``The paper explicitly models cross-expert redundancy by stacking experts and performing a joint Tucker factorization, instead of per-expert SVD/pruning.'' &
None \note{(the reviewer would become neutral)} \\
\addlinespace[3pt]
\rowcolor{palWarmTan!18}[0pt][0pt]
\multicolumn{2}{@{}l@{}}{\textbf{ICLR 2026 submission:} \textit{Temporal-Oriented Recipe}\qquad \textbf{Extracted:} \textcolor{palTerracotta}{Negative}\quad \textbf{Annotator:} \textcolor{palDeepTeal}{Positive}} \\
``The paper's contribution lies in the \sig{novel combination} and empirical validation of existing techniques \sig{rather than the introduction of a new method}.'' &
``Q-Formers, memory banks, MoE, and temporal captioning/grounding tasks are all \sig{established concepts}.'' \note{(still conveys the extracted negative polarity)} \par\vspace{3pt}
``the work could be perceived as a \sig{large-scale engineering effort} \dots \sig{rather than a paper presenting a fundamental new algorithm or architecture}.'' \note{(still conveys the extracted negative polarity)} \\
\bottomrule
\end{tabularx}
\caption[Signal extraction error examples]{\textbf{Signal extraction errors in context.} The three extracted signals whose polarity the annotator disputed (Appendix~\ref{subsub:novelty-sig-validation}), alongside the other novelty signals extracted from the same review. For \textit{Gym4ReaL} and \textit{Temporal-Oriented Recipe}, the reviewer's other signals carry the extracted polarity, so the error does not change the reviewer's overall verdict. For \textit{TD-MoE}, removing the signal leaves the reviewer neutral, which could drop the idea from the validated novel pool, but the reviewer does not contradict its novel label.}
\label{tab:signal_extraction_errors}
\end{table*}

\begin{promptfigure*}{Remove evaluation information prompt.}{prompt:remove-findings}
  \promptshortcaption{Remove Evaluation Information}
  \promptfile{Remove Evaluation Information Prompt}{prompts/remove_eval_data.jinja2}
\end{promptfigure*}

Finally, we apply \texttt{claude-opus-4-6} to the remaining ideas to remove evaluation information related to actual results or findings specified in the paper (unless this data is key to the central contribution). Prompt~\ref{prompt:remove-findings} presents the relevant prompt for this step.

\subsection{Benchmark Data Setups Construction} \label{sub:benchmark_setups}

\begin{table}[H]
\centering
\footnotesize
\setlength{\tabcolsep}{3pt}
\renewcommand{\arraystretch}{1.1}
\newcommand{\src}[1]{{\scriptsize\textcolor{black!55}{#1}}}
\begin{tabular}{@{}l cc cc@{}}
\toprule
 & \Dpos & \Dneg & Pairwise & Pointwise \\
\midrule
\humanonly & 154 & 145\,\src{ICLR} & 154 & 299 \\
\humangen  & 154 & 154\,\src{LLM}  & 154 & 308 \\
\bottomrule
\end{tabular}
\caption[Data statistics]{\textbf{Data statistics.} Both setups share the novel pool $\Dpos$ (validated ICLR ideas) and differ in the lower-novelty pool $\Dneg$: validated ICLR ideas in \humanonly, LLM-generated ideas in \humangen.}
\label{tab:benchmark_stats}
\end{table}

\begin{promptfigure*}{Idea generation prompt.}{prompt:prompted-llm-idea-gen}
  \promptshortcaption{Idea Generation}
  \promptfile{Idea Generation Prompt}{prompts/idea_generation.jinja2}
\end{promptfigure*}

\paragraph{\humanonly.} Ideas collected through the process described in Appendix~\ref{sub:human_data_collection} are used in the pointwise format as is, or paired according to their primary ICLR area (e.g., ``\textit{generative models}'') in the pairwise format. Since we have more high-novelty human data than low-novelty human data, a low-novelty idea might be paired with a high-novelty idea more than once.
\paragraph{\humangen.} We pair each high-novelty, human-authored idea with an LLM-generated counterpart designed to simulate standard research generation. To produce this baseline, we seed \texttt{claude-sonnet-4-5} with the human idea's ICLR area (e.g., ``\textit{generative models}''). The model is instructed to reason about potential research directions before proposing a final idea, using the prompt detailed in Prompt~\ref{prompt:prompted-llm-idea-gen}. We use identical generation parameters (\texttt{max\_tokens=8192}, default settings). As before, these pairs are decoupled to create the pointwise data.

\subsection{Expert Validation} \label{sub:expert_validation}

We ran a small blind study to check that the weak-labeling assumption behind \humangen (\S\ref{sub:idea_collection}) is not obviously violated. A computer science professor with broad expertise across AI-related domains evaluated 30 idea pairs, chosen to concentrate on the cases where the assumption is most likely to fail. Twenty-one of them are all the pairs that our top-performing pairwise judge (\texttt{claude-opus-4-6}) got wrong on \humangen under the reference configuration. Four more are sampled at random from its errors on a variant of \humangen whose \Dneg{} is generated by \texttt{gpt-5.4}, where the same judge errs far more often. The remaining five are controls that \texttt{claude-opus-4-6} got right. The expert saw each pair with the idea order shuffled and no indication of source, judge verdict, or which pairs were controls. The expert received the same instructions as our reference judges (Prompts~\ref{prompt:baseline-pairwise} and~\ref{prompt:baseline-pointwise}). For each pair, the expert was asked to (1) identify the more novel idea and (2) decide whether the idea they ranked lower is novel, providing written justifications and citing relevant prior literature when possible. The expert sided with the validated novel idea in all 30 pairs, including all 25 judge-error pairs, and judged none of the 30 generated ideas novel (approximately five hours of annotation in total). Table~\ref{tab:judge_error_examples} gives representative examples of these annotations.

\begin{table*}[t]
\centering
\footnotesize
\setlength{\tabcolsep}{4pt}
\renewcommand{\arraystretch}{1.0}
\begin{tabularx}{\textwidth}{@{}>{\raggedright\arraybackslash}p{0.13\textwidth} X@{}}
\toprule
\multicolumn{2}{@{}l}{\textit{ICLR area: applications to computer vision, audio, language, and other modalities}} \\
\addlinespace[2pt]
\textbf{0} \newline (validated novel) & Despite advances in pretraining with extended context sizes, large language models (LLMs) still face challenges in effectively utilizing real-world long-context information, primarily due to insufficient long-context alignment caused by data quality issues, training inefficiencies, and the lack of well-designed optimization objectives. To address these limitations, we propose a framework named Short-to-Long Preference Optimization (SoLoPO), decoupling long-context preference optimization (PO) into two components: short-context PO and short-to-long reward alignment (SoLo-RA), supported by both theoretical and empirical evidence. Specifically, short-context PO leverages preference pairs sampled from short contexts to enhance the model's contextual knowledge utilization ability. Meanwhile, SoLo-RA explicitly encourages reward score consistency for the responses when conditioned on both short and long contexts that contain identical task-relevant information. This facilitates transferring the model's ability to handle short contexts into long-context scenarios. SoLoPO is compatible with mainstream preference optimization algorithms, while substantially improving the efficiency of data construction and training processes. \\
\addlinespace[3pt]
\textbf{1} \newline (weakly labeled lower-novelty) & Multimodal understanding requires more than aligning representations across vision, audio, and language---it demands reasoning about temporal precedence and causal relationships between modalities. Existing approaches treat modalities symmetrically through attention mechanisms, failing to capture how information in one modality temporally influences or explains events in another. We introduce Causal Multimodal Transformers, a framework that explicitly models directional causal dependencies across asynchronous modality streams. Our architecture incorporates learned temporal offsets and causal masking patterns that respect the natural information flow between modalities, enabling the model to distinguish between coincidental co-occurrence and genuine causal influence. By decomposing cross-modal attention into temporally-ordered causal graphs, the framework learns which modality provides predictive information for events in others at different time scales. This causally-aware design enhances interpretability and enables counterfactual reasoning---answering questions like ``what would the visual scene be if this sound had not occurred?'' Applications span video understanding, audio-visual speech processing, and multimodal content generation where temporal causality is fundamental to meaning. \\
\addlinespace[3pt]
\textbf{Expert} \newline selected: 0 & ``The idea in 1 of causal/temporal dependence between modalities is old. Temporal offsets, causal masks with directional cross-modal attention etc. It's all a generic shallow combination of these things. e.g., https://arxiv.org/abs/1906.00295 (\textgreater{}3K citations). The idea in 0 of decoupling long-context preference optimization into short and short-to-long seems clever and original, and a very specific method invention.'' \\
\addlinespace[2pt]
\midrule
\addlinespace[2pt]
\multicolumn{2}{@{}l}{\textit{ICLR area: unsupervised, self-supervised, semi-supervised, and supervised representation learning}} \\
\addlinespace[2pt]
\textbf{0} \newline (weakly labeled lower-novelty) & Representation learning paradigms---unsupervised, self-supervised, semi-supervised, and supervised---are typically treated as distinct methodologies requiring separate architectural designs and training protocols. We introduce Spectrum Learning, a unified framework that views supervision as a continuous spectrum rather than discrete categories. Our approach employs a meta-learned weighting mechanism that dynamically modulates the contribution of multiple learning objectives based on local supervision density in the feature space. By treating each data point as existing along a supervision gradient, the framework adaptively combines contrastive self-supervised losses, consistency regularization, pseudo-labeling, and supervised classification within a single coherent optimization. A novel supervision-aware attention module enables the model to identify which learning paradigm is most informative for different regions of the data manifold. This fluid integration allows seamless transitions as supervision availability changes, from fully unsupervised to fully supervised settings, without architectural modification. Spectrum Learning naturally handles mixed supervision scenarios common in practice, where different samples have varying levels of annotation quality and granularity, providing a principled approach to leveraging all available learning signals simultaneously. \\
\addlinespace[3pt]
\textbf{1} \newline (validated novel) & Multi-view clustering integrates the consistency and complementarity of different views to achieve unsupervised data grouping. Existing multi-view clustering methods primarily confront two challenges: i) they generally perform feature extraction in the feature domain, which is sensitive to noise and may neglect cluster-specific information that is indistinguishable in the original space; ii) current dynamic fusion methods adopt static strategies to learn weights, lacking capability to adjust strategies adaptively under complex scenarios according to variations in data distribution and view quality. To address these issues, we propose a large language model assisted dynamic agent for multi-view clustering (LLM-DAMVC), a novel framework that recasts multi-view clustering as a dynamic decision-making problem orchestrated by a large language model. Specifically, each view is equipped with complementary agents dedicated to feature extraction. A dual-domain contrastive module is introduced to optimize feature consistency and enhance cluster separability in both the feature domain and frequency domain. Additionally, an LLM-assisted view fusion mechanism provides a flexible fusion weight learning strategy that can be adaptively applied to complex scenarios and significantly different views. \\
\addlinespace[3pt]
\textbf{Expert} \newline selected: 1 & ``1: multi view clustering, with view fusion with dynamic decisions controlled by an agent according to view quality, is interesting and not common (especially given the timeframe is 2025). 0: combining supervised losses, pseudo-labeling, consistency regularization, and unsupervised objectives, meta-learning sample weights... standard, generic and vague combination of standard things... https://arxiv.org/abs/1905.02249'' \\
\bottomrule
\end{tabularx}
\caption[Expert annotation examples]{\textbf{Expert annotation examples.} Representative pairs from the blind annotation study on \humangen, alongside the rater's written justification.}
\label{tab:judge_error_examples}
\end{table*}

\section{Additional Judge Details} \label{sec:judge_implementation_dits}

\begin{promptfigure*}{\textbf{Reference pairwise judge prompt}, shown exactly as sent to the judge apart from the two \slot{IDEA} placeholders, which hold the ideas under comparison.}{prompt:baseline-pairwise}
  \promptshortcaption{Reference pairwise judge}
  \renewcommand{\promptfontsize}{\promptfontsmall}%
  \promptskeletonfile{Reference Pairwise Judge Prompt}{prompts/judge_baseline_pairwise.txt}
\end{promptfigure*}

\begin{promptfigure*}{\textbf{Reference pointwise judge prompt.} The pointwise counterpart of Prompt~\ref{prompt:baseline-pairwise}.}{prompt:baseline-pointwise}
  \promptshortcaption{Reference pointwise judge}
  \renewcommand{\promptfontsize}{\promptfontsmall}%
  \promptskeletonfile{Reference Pointwise Judge Prompt}{prompts/judge_baseline_pointwise.txt}
\end{promptfigure*}

\subsection{Prompted Judges}\label{sub:prompted_judges}
All prompted judge models use \texttt{max\_tokens=8192}. Additionally, all models (except \texttt{claude-sonnet-4-5}, which does not support the option) are configured with \texttt{reasoning\_effort="high"}, using default values for all other parameters. For the pointwise evaluation, we call the judge three times and take the majority vote. For the pairwise evaluation, we mitigate position bias by adopting the method from \citet{wang2023largelanguagemodelsfair} and calling the judge three times per presentation order ($A$ vs.\ $B$ and $B$ vs.\ $A$). When a judge selects an idea as the winner, that idea is assigned a score of 1, while the losing idea receives a 0. The final winner is determined by the higher average score, and if both ideas receive the same average, the match is declared a tie. Prompts~\ref{prompt:baseline-pairwise} and~\ref{prompt:baseline-pointwise} give the reference pairwise and pointwise prompts, respectively.

\subsection{Strong Retrieval Judge} \label{sub:strong_retrieval_judge}
\begin{promptfigure*}{\textbf{Strong retrieval judge prompt.} The judge searches the literature itself instead of being handed a fixed candidate list.}{prompt:strong-retrieval-judge}
  \promptshortcaption{Strong retrieval judge}
  \renewcommand{\promptfontsize}{\promptfontsmall}%
  \promptskeletonfile{Strong Retrieval Judge Prompt}{prompts/web_search_prompt.txt}
\end{promptfigure*}

The strong retrieval judge introduced in \S\ref{subsec:retrieval_limitations} is built on \texttt{gpt-5.6-sol}. We set \texttt{reasoning\_effort} to \texttt{xhigh} and enable the provider-hosted \texttt{web\_search} tool, restricted to \texttt{arxiv.org}. We impose no cap on the number of tokens used, so the judge is free to issue as many searches and inspect as many results as it deems necessary before committing to a verdict. Given its higher price, we call the judge once per idea, without verdict aggregation. The prompt for this judge is given in Prompt~\ref{prompt:strong-retrieval-judge}. It is identical to the reference prompt (Prompt~\ref{prompt:baseline-pointwise}), except for additional search-tool-related instructions. The added guidance constrains the search to papers published before the retrieval cutoff (2025-03-01) and instructs the judge to disregard any hit that is the idea's own preprint, mirroring a human reviewer who happens to stumble upon the paper under review.

\begin{figure}[t]
\centering
\includegraphics[width=0.9\columnwidth]{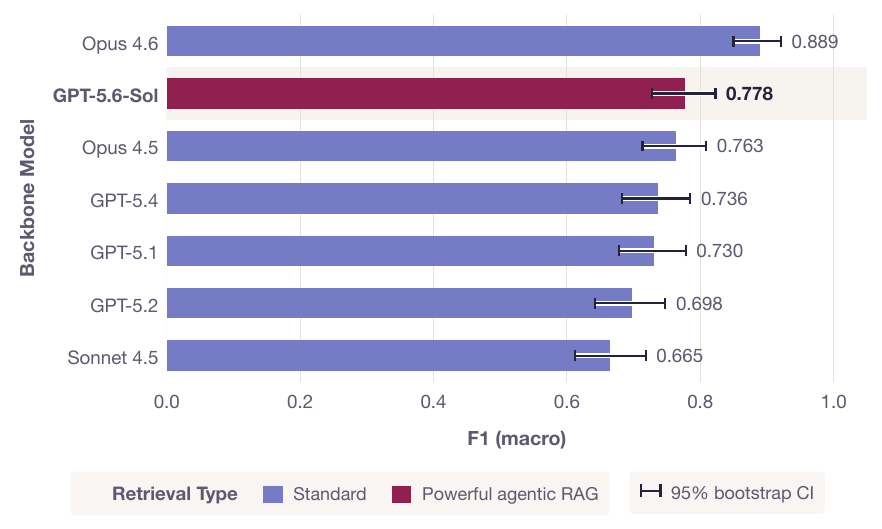}
\caption[Stronger retrieval judge]{\textbf{Stronger retrieval doesn't help.} Pointwise F1 on \humangen (without verdict aggregation). A strong agentic RAG judge (\texttt{gpt-5.6-sol} with iterative web search) shows no significant gain over judges with standard retrieval (\S\ref{sec:experimental_setup}), despite a newer backbone, more reasoning, and unlimited search.}
\label{fig:retrieval_faceoff}
\end{figure}

\subsection{Dedicated Novelty Judges} \label{sub:existing_evaluators}
We follow the implementations released by \citet{Shahid2025LiteratureGroundedNA}\footnote{\url{https://github.com/simra-shahid/idea_novelty_checker}} for both judges. Each is evaluated with \texttt{claude-opus-4-6} and \texttt{gpt-5.4} as its backbone, at \texttt{reasoning\_effort} \texttt{"high"} and a single call per decision, without verdict aggregation. Unlike our prompted judges, both perform their own Semantic Scholar retrieval rather than receiving a fixed candidate set. We restrict those searches to papers published before the retrieval cutoff (2025-03-01) and discard any hit whose title matches the paper the idea was drawn from.

We use the following configurations for the judges:
\begin{itemize}
  \item \textbf{Idea Novelty Checker}. We keep the upstream defaults: keyword, title, and snippet search for paper collection, Specter2 embedding filtering, a RankGPT rerank in its \texttt{priority} variant, and a verdict conditioned on the top $10$ ranked papers, together with the relaxed in-context example set. The rerank is run with \texttt{gpt-5.4-mini} under both backbones.
  \item \textbf{AI-Scientist}. We allow the agent up to $10$ search rounds, the upstream default, feeding each round's results into the next round's prompt. 
\end{itemize}

\begin{figure}[!htb]
    \centering
    \includegraphics[width=\columnwidth]{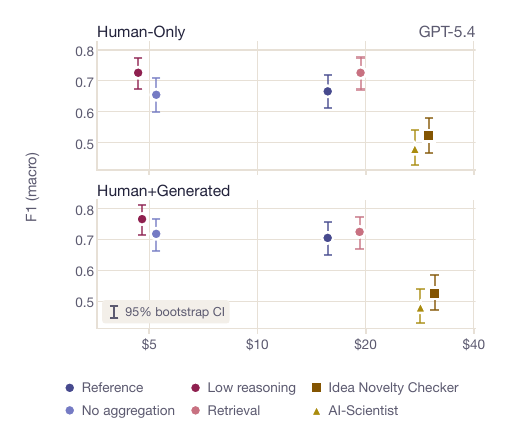}
    \caption[Cost vs.\ macro-F1 (\texttt{gpt-5.4})]{\textbf{Cost vs.\ pointwise macro-F1} for \texttt{gpt-5.4}, the counterpart of Figure~\ref{fig:pareto_pointwise_opus}, comparing prompted-judge configurations against the dedicated novelty judges. The same pattern holds: every prompted configuration outscores both dedicated judges at a lower cost.}
    \label{fig:pareto_pointwise_gpt}
\end{figure}

\begin{table*}[t]
\centering
\footnotesize
\setlength{\tabcolsep}{4pt}
\begin{tabular*}{\textwidth}{@{\extracolsep{\fill}}llccccr}
\toprule
\textbf{Evaluator} & \textbf{Judge} & \textbf{$k$} & \textbf{F$_1$} & \textbf{F$_1^{+}$} & \textbf{F$_1^{-}$} & \textbf{Cost (USD)} \\
\midrule
\multicolumn{7}{l}{\textit{\humanonly}} \\
AI-Scientist & \texttt{claude-opus-4-6} & 1 & $0.652_{\pm0.055}$ & $0.743_{\pm0.050}$ & $0.561_{\pm0.079}$ & 77.91 \\
AI-Scientist & \texttt{gpt-5.4} & 1 & $0.481_{\pm0.056}$ & $0.293_{\pm0.084}$ & $0.668_{\pm0.053}$ & 27.41 \\
Idea Novelty Checker & \texttt{claude-opus-4-6} & 1 & $0.560_{\pm0.059}$ & $0.711_{\pm0.050}$ & $0.408_{\pm0.087}$ & 30.11 \\
Idea Novelty Checker & \texttt{gpt-5.4} & 1 & $0.521_{\pm0.057}$ & $0.701_{\pm0.050}$ & $0.342_{\pm0.088}$ & 29.98 \\
Prompted (no aggregation) & \texttt{claude-opus-4-6} & 1 & $0.782_{\pm0.047}$ & $0.792_{\pm0.051}$ & $0.772_{\pm0.054}$ & 4.56 \\
Prompted (no aggregation) & \texttt{gpt-5.4} & 1 & $0.655_{\pm0.055}$ & $0.578_{\pm0.076}$ & $0.732_{\pm0.051}$ & 5.23 \\
Prompted (reference) & \texttt{claude-opus-4-6} & 3 & $\mathbf{0.806}_{\pm0.046}$ & $\mathbf{0.815}_{\pm0.048}$ & $\mathbf{0.796}_{\pm0.052}$ & 13.67 \\
Prompted (reference) & \texttt{gpt-5.4} & 3 & $0.666_{\pm0.053}$ & $0.592_{\pm0.076}$ & $0.740_{\pm0.050}$ & 15.68 \\
Prompted (low reasoning) & \texttt{claude-opus-4-6} & 3 & $0.739_{\pm0.050}$ & $0.729_{\pm0.059}$ & $0.748_{\pm0.053}$ & 2.43 \\
Prompted (low reasoning) & \texttt{gpt-5.4} & 3 & $0.726_{\pm0.051}$ & $0.685_{\pm0.066}$ & $0.767_{\pm0.049}$ & 4.66 \\
Prompted (retrieval) & \texttt{claude-opus-4-6} & 3 & $0.802_{\pm0.046}$ & $0.813_{\pm0.048}$ & $0.792_{\pm0.051}$ & 24.84 \\
Prompted (retrieval) & \texttt{gpt-5.4} & 3 & $0.726_{\pm0.052}$ & $0.685_{\pm0.067}$ & $0.767_{\pm0.050}$ & 19.36 \\
\addlinespace
\multicolumn{7}{l}{\textit{\humangen}} \\
AI-Scientist & \texttt{claude-opus-4-6} & 1 & $0.772_{\pm0.047}$ & $0.801_{\pm0.047}$ & $0.743_{\pm0.059}$ & 82.05 \\
AI-Scientist & \texttt{gpt-5.4} & 1 & $0.481_{\pm0.055}$ & $0.260_{\pm0.085}$ & $0.702_{\pm0.049}$ & 28.41 \\
Idea Novelty Checker & \texttt{claude-opus-4-6} & 1 & $0.642_{\pm0.055}$ & $0.728_{\pm0.050}$ & $0.556_{\pm0.077}$ & 30.66 \\
Idea Novelty Checker & \texttt{gpt-5.4} & 1 & $0.526_{\pm0.057}$ & $0.688_{\pm0.051}$ & $0.365_{\pm0.085}$ & 31.10 \\
Prompted (no aggregation) & \texttt{claude-opus-4-6} & 1 & $\mathbf{0.902}_{\pm0.034}$ & $\mathbf{0.895}_{\pm0.039}$ & $\mathbf{0.909}_{\pm0.033}$ & 4.91 \\
Prompted (no aggregation) & \texttt{gpt-5.4} & 1 & $0.718_{\pm0.052}$ & $0.646_{\pm0.073}$ & $0.791_{\pm0.045}$ & 5.23 \\
Prompted (reference) & \texttt{claude-opus-4-6} & 3 & $0.889_{\pm0.036}$ & $0.879_{\pm0.041}$ & $0.898_{\pm0.035}$ & 14.74 \\
Prompted (reference) & \texttt{gpt-5.4} & 3 & $0.705_{\pm0.053}$ & $0.625_{\pm0.075}$ & $0.786_{\pm0.045}$ & 15.68 \\
Prompted (low reasoning) & \texttt{claude-opus-4-6} & 3 & $0.854_{\pm0.040}$ & $0.835_{\pm0.049}$ & $0.874_{\pm0.037}$ & 2.59 \\
Prompted (low reasoning) & \texttt{gpt-5.4} & 3 & $0.766_{\pm0.049}$ & $0.718_{\pm0.064}$ & $0.814_{\pm0.044}$ & 4.77 \\
Prompted (retrieval) & \texttt{claude-opus-4-6} & 3 & $0.892_{\pm0.035}$ & $0.883_{\pm0.041}$ & $0.901_{\pm0.034}$ & 25.19 \\
Prompted (retrieval) & \texttt{gpt-5.4} & 3 & $0.725_{\pm0.052}$ & $0.667_{\pm0.069}$ & $0.783_{\pm0.047}$ & 19.23 \\
\bottomrule
\end{tabular*}
\caption[Cost-efficiency results]{\textbf{Pointwise results and cost} for every judge configuration, in the \humanonly (\emph{top}) and \humangen (\emph{bottom}) setups. $k$ is the number of judge calls per decision, F$_1$ is macro-F1, and F$_1^{+}$ and F$_1^{-}$ are the per-class scores on novel and non-novel ideas. Cost is the total USD spent judging the full set. Best macro-F1 per setup in \textbf{bold}. Subscripts are the half-width of the 95\% bootstrap confidence interval over test instances, from 10,000 resamples. These numbers underlie Figures~\ref{fig:pareto_pointwise_opus} and~\ref{fig:pareto_pointwise_gpt}.}
\label{tab:cost_pointwise}
\end{table*}

\paragraph{Additional results.} Table~\ref{tab:cost_pointwise} reports the full per-judge numbers behind Figure~\ref{fig:pareto_pointwise_opus}, and Figure~\ref{fig:pareto_pointwise_gpt} gives the corresponding cost-quality view for \texttt{gpt-5.4}. Under both backbones and in both setups, the dedicated judges are the most expensive configurations we run and the least accurate, i.e., they are not cost-effective.

\section{Controlled Evaluation Study Details}\label{sub:study_details}
\subsection{Implementation Details} \label{sub:study_implementation_details}

\begin{table*}[tp]
\centering
\footnotesize
\setlength{\tabcolsep}{2.5pt}
\renewcommand{\arraystretch}{1.15}
\newcommand{\yes}{{\scriptsize\faCheck}}
\newcommand{\no}{\textcolor{gray}{--}}
\newcommand{\rf}{\textcolor{gray}{\scriptsize ref}}
\begin{tabularx}{\textwidth}{@{}l Y{1} cc cc c@{}}
\toprule
 & & \multicolumn{2}{c}{\scriptsize\textit{Format}} & \multicolumn{2}{c}{\scriptsize\textit{Data setup}} & \\
\cmidrule(lr){3-4} \cmidrule(lr){5-6}
\textbf{Change} & \textbf{Description} & {\scriptsize Pt} & {\scriptsize Pw} & {\scriptsize HO} & {\scriptsize HG} & \textbf{Fig.} \\
\midrule
\multicolumn{7}{@{}l@{}}{\rule{0pt}{2.6ex}\textbf{Judge prompt}\ \ \textcolor{gray}{\textit{(reference: requests judges to assess ideas against novelty criteria defining what counts as novel)}}} \\
\addlinespace[2pt]
P1: Criteria without guardrails & Weaker criteria, w/o guardrails (e.g., no trivial combinations) & \yes & \yes & \yes & \yes & \ref{fig:ablation} \\
P2: Criteria removed & Drop novelty criteria entirely & \yes & \yes & \yes & \yes & \ref{fig:ablation} \\
P3: ``Which was judged novel?'' & Prior review found exactly one idea novel; judge predicts which & \no & \yes & \yes & \yes & \ref{fig:ablation_pairwise} \\
P4: ``Which is novel?'' & P3 minus the presumption of prior review & \no & \yes & \yes & \yes & \ref{fig:ablation_pairwise} \\
P5: ``Which was judged more novel?'' & Prior review found one idea \textit{more} novel; judge predicts which & \no & \yes & \yes & \yes & \ref{fig:ablation_pairwise} \\
\midrule
\multicolumn{7}{@{}l@{}}{\rule{0pt}{2.6ex}\textbf{Evaluation setup}\ \ \textcolor{gray}{\textit{(reference: no retrieval; \texttt{high} reasoning effort; free-text abstracts; verdict aggregation; $\Dneg$ by \texttt{sonnet-4-5})}}} \\
\addlinespace[2pt]
$+$ Related-work (retrieval) & Judges receive 5 abstracts of related work & \yes & \yes & \yes & \yes & \ref{fig:ablation} \\
Reasoning effort $=$ \texttt{low} & Reduced reasoning budget & \yes & \yes & \yes & \yes & \ref{fig:ablation} \\
Idea format & A two-field structured plan (Purpose $\mid$ Mechanism) & \yes & \yes & \yes & \yes & \ref{fig:ablation} \\
No verdict aggregation & Query pointwise judges once, pairwise once per direction & \yes & \yes & \yes & \yes & \ref{fig:ablation} \\
$\Dneg$ source & $\Dneg$ regenerated by \texttt{opus-4-5}, \texttt{gpt-5}, \texttt{gpt-5.1}, \texttt{gpt-5.4} & \yes & \yes & \no & \yes & \ref{fig:negatives_source} \\
\midrule
\multicolumn{7}{@{}l@{}}{\rule{0pt}{2.6ex}\textbf{Judge backbones}} \\
\addlinespace[2pt]
\multicolumn{7}{@{}l@{}}{\texttt{gpt-5.1}, \texttt{gpt-5.2}, \texttt{gpt-5.4}, \texttt{claude-sonnet-4-5}, \texttt{claude-opus-4-5}, \texttt{claude-opus-4-6}} \\
\bottomrule
\end{tabularx}
\caption[Controlled changes summary]{\textbf{Controlled changes summary.} Each change (\S\ref{sub:eval_dimensions}) modifies a single design choice relative to the reference configuration (\S\ref{sub:reference_config}), holding the rest of the novelty evaluation process fixed. Columns mark the formats (Pt: pointwise, Pw: pairwise) and data setups (HO: \humanonly, HG: \humangen) each change is run under. Beyond these single changes, we compare prompted judges to dedicated novelty judges (Fig.~\ref{fig:pareto_pointwise_opus}) and to a strong agentic RAG judge (Fig.~\ref{fig:retrieval_faceoff}).}
\label{tab:controlled_changes}
\end{table*}

\begin{promptfigure}{\textbf{P1: Criteria without guardrails, diff against Prompts~\ref{prompt:baseline-pairwise} and~\ref{prompt:baseline-pointwise}.} P1 deletes the three paragraphs shown, which constrain what counts as novel.}{prompt:diff-vague}
  \promptshortcaption{P1: Criteria without guardrails (diff)}
  \renewcommand{\promptfontsize}{\promptfontsmall}%
  \promptfile{Deleted from Both Reference Prompts}{prompts/diff_vague_criterion.txt}
\end{promptfigure}

\begin{promptfigure*}{\textbf{P2: Criteria removed, pairwise.} }{prompt:no-criterion-pairwise}
  \promptshortcaption{P2: Criteria removed (pairwise)}
  \renewcommand{\promptfontsize}{\promptfontsmall}%
  \promptskeletonfile{P2 Pairwise Judge Prompt}{prompts/no_criterion_pairwise.txt}
\end{promptfigure*}

\begin{promptfigure*}{\textbf{P2: Criteria removed, pointwise.} }{prompt:no-criterion-pointwise}
  \promptshortcaption{P2: Criteria removed (pointwise)}
  \renewcommand{\promptfontsize}{\promptfontsmall}%
  \promptskeletonfile{P2 Pointwise Judge Prompt}{prompts/no_criterion_pointwise.txt}
\end{promptfigure*}

\paragraph{Judge prompt.} Five variants replace the reference prompts (Prompts~\ref{prompt:baseline-pairwise} and~\ref{prompt:baseline-pointwise}): P1 and P2 in both the pairwise and the pointwise format, P3--P5 in the pairwise format only. \emph{P1: Criteria without guardrails} strips the novelty criteria of their guardrails against trivial novelty, deleting the three paragraphs shown in Prompt~\ref{prompt:diff-vague}. \emph{P2: Criteria removed} goes further and drops the criteria altogether (Prompts~\ref{prompt:no-criterion-pairwise} and~\ref{prompt:no-criterion-pointwise}). P3--P5 are not edits of the reference prompt but a separate lineage, adapted from \citet{Si2024CanLG}, whose original prompt asks which of two abstracts was accepted at a conference. \emph{P3: ``Which was judged novel?''} presumes that a prior review found exactly one of the two ideas novel and asks the judge to predict which (Prompt~\ref{prompt:p3-judged-novel}), \emph{P4: ``Which is novel?''} drops that presumption (Prompt~\ref{prompt:p4-is-novel}), and \emph{P5: ``Which was judged more novel?''} restores it in a comparative form (Prompt~\ref{prompt:p5-more-novel}).

\begin{promptfigure*}{\textbf{P3: ``Which was judged novel?''}, adapted from \citet{Si2024CanLG}, whose original prompt asks which of two abstracts was accepted at a conference.}{prompt:p3-judged-novel}
  \promptshortcaption{P3: ``Which was judged novel?''}
  \promptfile{P3 Judge Prompt (\citealp{Si2024CanLG}, Modified for Novelty Evaluation)}{prompts/si_pairwise_baseline.jinja2}
\end{promptfigure*}

\begin{promptfigure*}{\textbf{P4: ``Which is novel?''}, which drops P3's reviewer and venue framing.}{prompt:p4-is-novel}
  \promptshortcaption{P4: ``Which is novel?''}
  \promptfile{P4 Judge Prompt}{prompts/si_pairwise_no_review.jinja2}
\end{promptfigure*}

\begin{promptfigure*}{\textbf{P5: ``Which was judged more novel?''}, which replaces P3's binary novel/not-novel contrast with a graded one.}{prompt:p5-more-novel}
  \promptshortcaption{P5: ``Which was judged more novel?''}
  \promptfile{P5 Judge Prompt}{prompts/si_pairwise_comparative.jinja2}
\end{promptfigure*}

\paragraph{Idea format.}
\begin{promptfigure*}{Research plan extraction prompt.}{prompt:research-plan-extraction}
  \promptshortcaption{Research plan extraction}
  \promptfile{Research Plan Extraction Prompt}{prompts/extract_research_plan.jinja2}
\end{promptfigure*}
We use \texttt{claude-opus-4-6} to extract a two-field structured plan from the free-form idea, following the prompt detailed in Prompt~\ref{prompt:research-plan-extraction}. We then treat the resulting plans as the ideas, and give them to the judges as described in Appendix~\ref{sub:prompted_judges}.

\paragraph{Retrieval.}
\begin{promptfigure*}{Retrieval contribution prompt.}{prompt:retrieval-contribution}
  \promptshortcaption{Retrieval contribution}
  \promptfile{Retrieval Contribution Prompt}{prompts/retrieval_contribution.jinja2}
\end{promptfigure*}

\begin{promptfigure*}{Retrieval query generation prompt.}{prompt:retrieval-query-generation}
  \promptshortcaption{Retrieval query generation}
  \promptfile{Retrieval Query Generation Prompt}{prompts/retrieval_query_generation.jinja2}
\end{promptfigure*}

We conduct retrieval for each benchmark idea by first extracting up to three main contributions using \texttt{claude-opus-4-6} with the prompt specified in Prompt~\ref{prompt:retrieval-contribution}. Following this extraction, we instruct the model to simulate the thought process of a novelty reviewer and generate five targeted queries designed to probe the novelty of these specific contributions, utilizing the prompt detailed in Prompt~\ref{prompt:retrieval-query-generation}. We then leverage the queries to conduct a Paper-Finder\footnote{\url{https://github.com/allenai/asta-paper-finder}} search and retrieve initial candidates. Next, we filter out the ones published after the retrieval cutoff (2025-03-01). The cutoff sits about six months before the ICLR 2026 abstract deadline (2025-09-19). Authors often post drafts to arXiv before submitting, and this margin keeps such preprints, which would reveal the very idea under evaluation, out of the retrieved work. As an additional safeguard, we also discard the judged idea itself if it surfaces among the candidates. For each query, we keep the remaining candidate with the highest relevance score, yielding $5$ abstracts as the final retrieved related work. 
Prompt~\ref{prompt:diff-retrieval} shows how those abstracts enter the judge prompt, and the accompanying rewording of the novelty criteria, which ask about originality relative to the supplied prior work. We show the pairwise criteria; the pointwise criteria are reworded similarly.

\begin{promptfigure*}{\textbf{Retrieval, diff against Prompts~\ref{prompt:baseline-pairwise} and~\ref{prompt:baseline-pointwise}.} \emph{Top}: the retrieved abstracts are appended to each idea inside a delimited block, shown here for the pairwise reference prompt; the pointwise reference prompt takes a single such block, with the delimiters reading \texttt{[Start Related Work]} and \texttt{[End Related Work]}. \emph{Bottom}: the criteria pick up the three highlighted insertions, so that they ask about originality relative to the supplied prior work. The elided paragraphs and the rest of the prompt are unchanged.}{prompt:diff-retrieval}
  \promptshortcaption{Diff: retrieval}
  \renewcommand{\promptfontsize}{\promptfontsmall}%
  \promptskeletonfile{Idea Blocks, With Related Work Appended}{prompts/diff_retrieval_ideas.txt}
  \vspace{2mm}
  \promptdifffile{Criteria, Reworded}{prompts/diff_retrieval_criterion.txt}
\end{promptfigure*}

\paragraph{Significance testing.}
We test the significance of every change reported relative to the reference configuration (\S\ref{sub:reference_config}) with a two-sided paired non-parametric bootstrap of $100{,}000$ resamples. For a given judge and metric, we resample the evaluation instances with replacement, apply the same resampled indices to the predictions of the reference and of the modified configuration, and recompute the difference between the two. The bootstrap is corpus-level: on each resample we recompute the metric (macro-F1, soft-accuracy, or strict-accuracy) over the full resampled set. From the resulting distribution of differences we build a bias-corrected and accelerated (BCa) $95\%$ confidence interval, and mark a difference with * when this interval excludes zero (i.e., $p < 0.05$).

\subsection{Additional Results} \label{sub_additional_ablation_results}
\begin{figure*}[!htb]
    \centering
    \includegraphics[width=\textwidth]{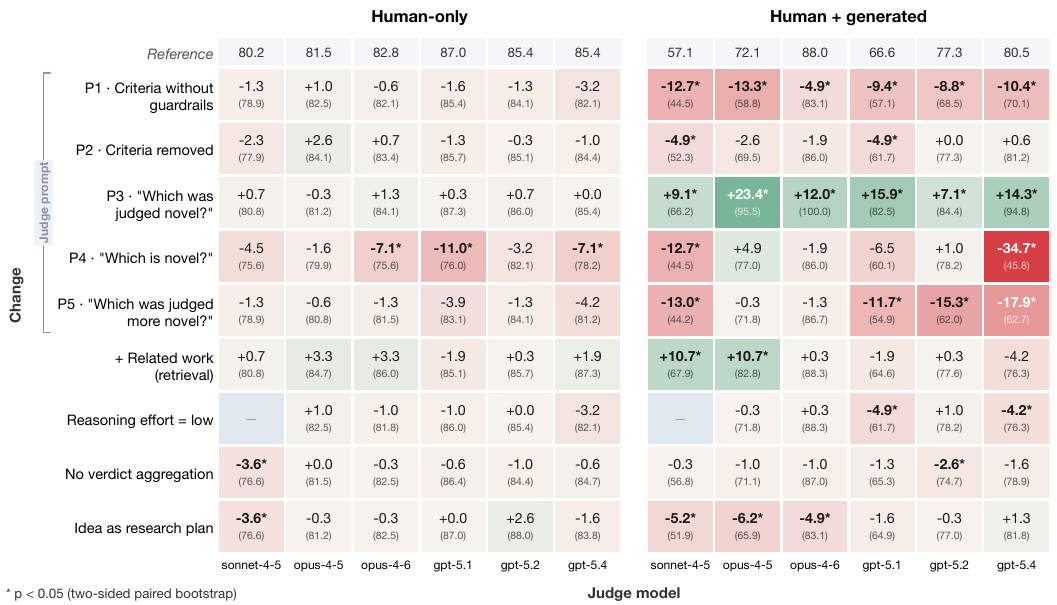}
    \caption[Pairwise soft-accuracy results]{Pairwise controlled-study results under \textit{soft}-accuracy, where ties count as half-correct. Each cell reports the change relative to the reference configuration (top row), with the resulting absolute soft-accuracy in parentheses. Asterisks mark statistically significant changes.}\label{fig:ablation_pairwise_soft}
\end{figure*}

\paragraph{Soft accuracy results.} Figure~\ref{fig:ablation_pairwise_soft} repeats the pairwise experiment in Figure~\ref{fig:ablation_pairwise} under \emph{soft} accuracy, where ties count as half-correct. Judges obtain higher results under the soft metric, but the effects of the controlled changes follow the same patterns, with high variance and with the \humanonly data being less sensitive to perturbations.

\begin{figure*}[!htb]
    \centering
    \begin{subfigure}{\textwidth}
        \centering
        \includegraphics[width=\textwidth]{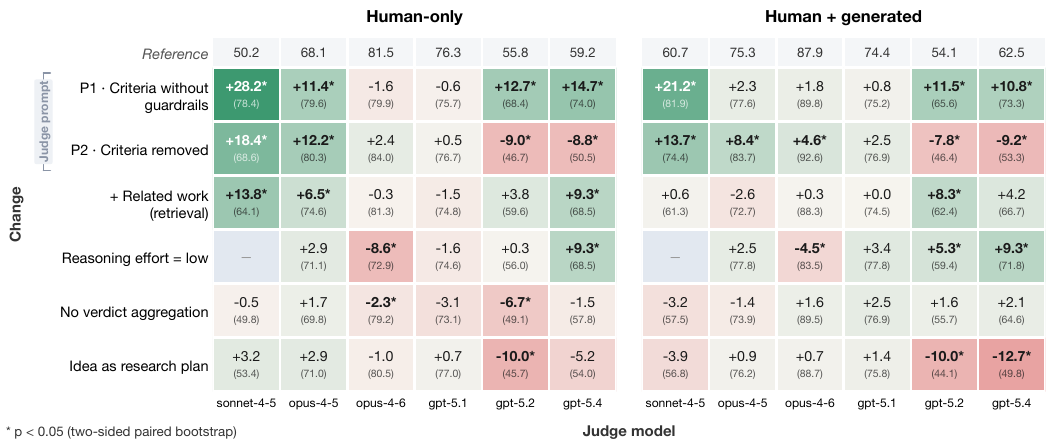}
        \caption{F1 on the \textbf{novel} class.}\label{fig:ablation_pointwise_f1_pos}
    \end{subfigure}

    \vspace{1em}

    \begin{subfigure}{\textwidth}
        \centering
        \includegraphics[width=\textwidth]{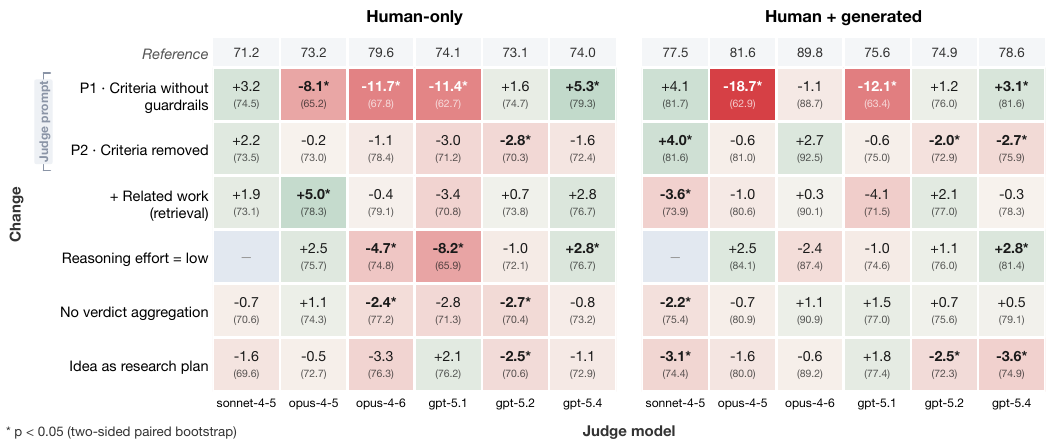}
        \caption{F1 on the \textbf{not-novel} class.}\label{fig:ablation_pointwise_f1_neg}
    \end{subfigure}
    \caption[Per-class pointwise results]{Per-class decomposition of the pointwise controlled-study results. Each cell reports the change relative to the reference configuration (top row), with the resulting absolute F1 in parentheses. Asterisks mark statistically significant changes.}\label{fig:ablation_pointwise_f1_per_class}
\end{figure*}

\paragraph{Per-class pointwise results.} Figure~\ref{fig:ablation_pointwise_f1_per_class} reports the controlled-study results using F1 on the \emph{novel} class (Figure~\ref{fig:ablation_pointwise_f1_pos}) and F1 on the \emph{not-novel} class (Figure~\ref{fig:ablation_pointwise_f1_neg}).

\begin{figure*}[!htb]
    \centering
    \includegraphics[width=\textwidth]{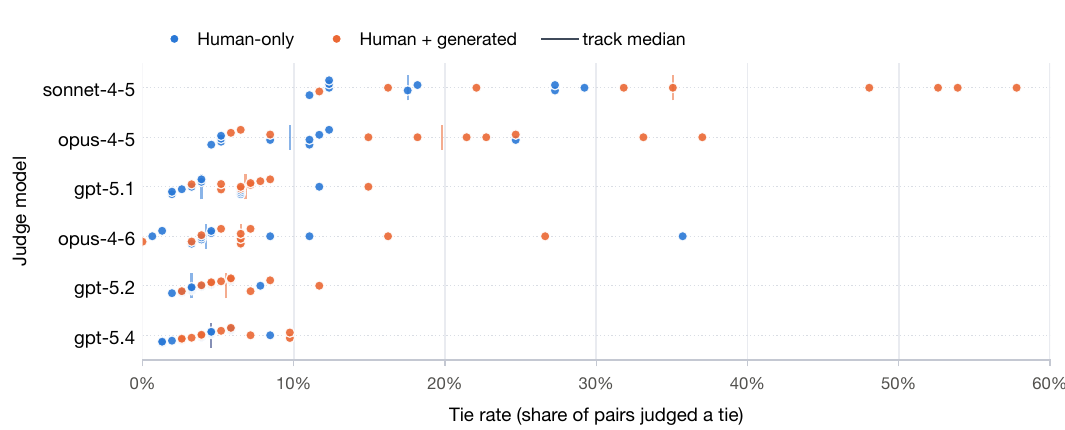}
    \caption[Tie rates across configurations]{\textbf{Tie rates across controlled configurations}. Each point is one configuration of Figure~\ref{fig:ablation_pairwise} for a given judge (roughly $20$ data points per judge), with the vertical bar marking the median per setup. Every judge ties more often in the \humangen setup than in the \humanonly one. Exact per-configuration values are in Table~\ref{tab:tie_rates}.}\label{fig:tie_rates}
\end{figure*}

\paragraph{Tie rates.} Figure~\ref{fig:tie_rates} reports the tie rate of each pairwise judge across the configurations presented in Figure~\ref{fig:ablation_pairwise} (roughly $20$ data points per judge). Table~\ref{tab:tie_rates} shows the exact tie rate per configuration and model. We observe that judges tie more on average once $\Dneg$ is LLM-generated, with \texttt{claude-sonnet-4-5} tying on more than half of the pairs in certain configurations. Additionally, verdict aggregation substantially reduces ties.

\begin{table*}[!htb]
\centering
\footnotesize
\setlength{\tabcolsep}{3pt}
\begin{tabular}{@{}l cccccc@{\hspace{8pt}}cccccc@{}}
\toprule
& \multicolumn{6}{c}{\textbf{\humanonly}} & \multicolumn{6}{c}{\textbf{\humangen}} \\
\cmidrule(lr){2-7} \cmidrule(lr){8-13}
\textbf{Configuration} & son-4-5 & op-4-5 & op-4-6 & 5.1 & 5.2 & 5.4 & son-4-5 & op-4-5 & op-4-6 & 5.1 & 5.2 & 5.4 \\
\midrule
Reference                                   & 12.3 &  4.5 &  0.6 &  6.5 &  3.2 &  4.5 & 35.1 & 18.2 &  3.2 &  7.1 &  2.6 &  5.2 \\
P1 $\cdot$ Criteria without guardrails      & 29.2 &  5.2 &  3.2 &  1.9 &  3.2 &  1.9 & 52.6 & 21.4 &  5.2 &  7.8 &  8.4 &  3.9 \\
P2 $\cdot$ Criteria removed                 & 27.3 & 11.0 &  4.5 &  2.6 &  2.6 &  3.9 & 57.8 & 24.7 &  7.1 &  6.5 &  3.9 &  2.6 \\
P3 $\cdot$ ``Which was judged novel?''      & 12.3 & 11.7 & 11.0 &  1.9 &  4.5 &  4.5 & 11.7 &  6.5 &  0.0 &  6.5 &  5.2 &  2.6 \\
P4 $\cdot$ ``Which is novel?''              & 12.3 & 24.7 & 35.7 &  6.5 &  5.8 &  4.5 & 16.2 & 37.0 & 26.6 &  7.1 &  4.5 &  9.7 \\
P5 $\cdot$ ``Which was judged more novel?'' & 11.0 & 11.0 &  8.4 &  3.9 &  1.9 &  1.3 & 22.1 & 22.7 & 16.2 &  8.4 &  5.8 &  5.8 \\
+ Related work (retrieval)                  & 17.5 &  8.4 &  4.5 &  6.5 &  3.9 &  5.8 & 31.8 &  8.4 &  3.9 &  3.2 &  5.8 &  7.1 \\
Reasoning effort = low                      &      &  5.2 &  3.9 &  3.2 &  3.2 &  5.8 &      &  5.8 &  6.5 &  5.2 &  4.5 &  3.2 \\
No verdict aggregation                      & 27.3 & 12.3 &  1.3 & 11.7 &  7.8 &  8.4 & 53.9 & 33.1 &  6.5 & 14.9 & 11.7 &  9.7 \\
Idea as research plan                       & 18.2 &  5.2 &  3.9 &  3.9 &  3.2 &  1.3 & 48.1 & 14.9 &  6.5 &  5.2 &  7.1 &  3.9 \\
\midrule
\textit{Mean}                               & 18.6 &  9.9 &  7.7 &  4.9 &  4.0 &  4.2 & 36.6 & 19.3 &  8.2 &  7.2 &  6.0 &  5.4 \\
\bottomrule
\end{tabular}
\caption[Tie rates per controlled configuration]{\textbf{Tie rates (\%) across controlled configurations}, the numbers behind Figure~\ref{fig:tie_rates}. Judges are abbreviated: \texttt{son-4-5} is \texttt{claude-sonnet-4-5}, \texttt{op-4-5} is \texttt{claude-opus-4-5}, \texttt{op-4-6} is \texttt{claude-opus-4-6}, \texttt{5.1} is \texttt{gpt-5.1}, \texttt{5.2} is \texttt{gpt-5.2}, \texttt{5.4} is \texttt{gpt-5.4}. Judges are substantially more uncertain on the \humangen set.}
\label{tab:tie_rates}
\end{table*}

\begin{figure*}[!htb]
    \centering
    \begin{subfigure}{\textwidth}
        \centering
        \includegraphics[width=\textwidth]{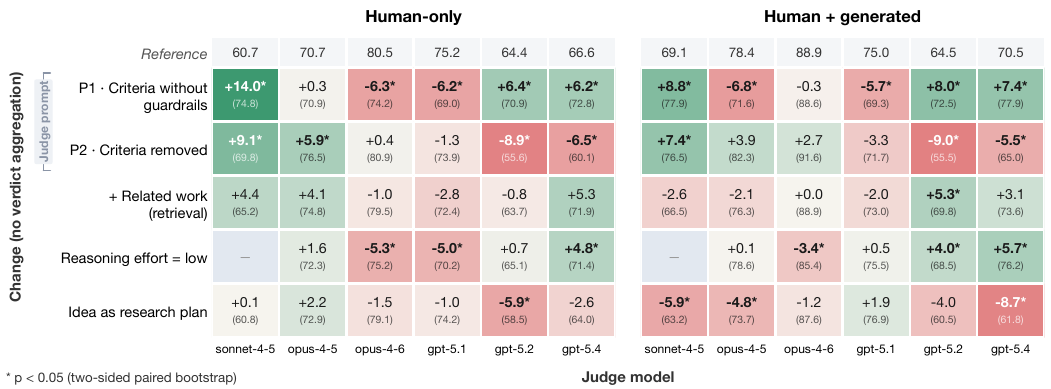}
        \caption{\textbf{Pointwise} judging, macro-F1.}\label{fig:ablation_pointwise_k1}
    \end{subfigure}

    \begin{subfigure}{\textwidth}
        \centering
        \includegraphics[width=\textwidth]{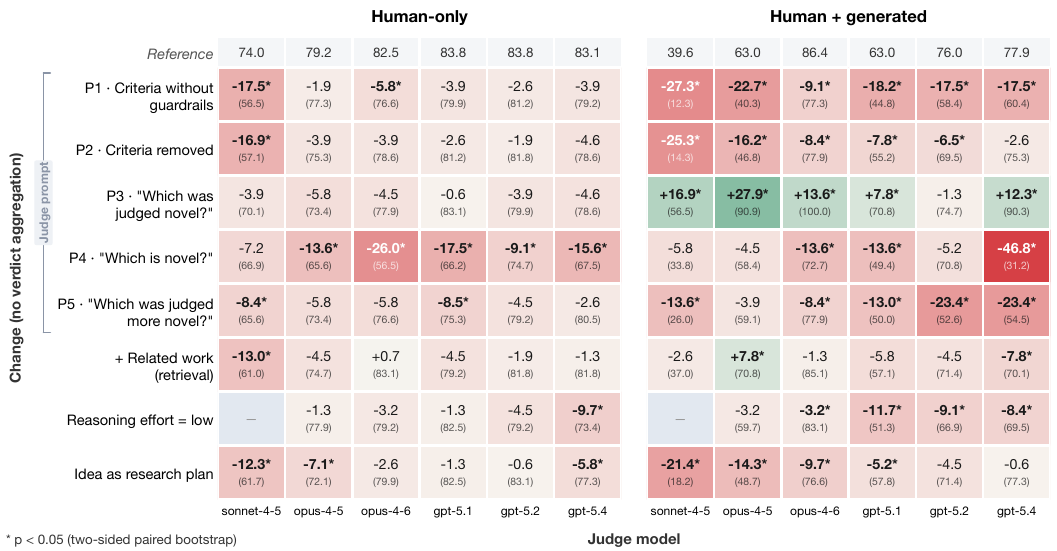}
        \caption{\textbf{Pairwise} judging, strict-accuracy (ties count as failures).}\label{fig:ablation_pairwise_k1}
    \end{subfigure}
    \caption[Results without verdict aggregation]{\textbf{Results without verdict aggregation}. Cells report the change from the reference configuration, with the absolute value in parentheses. The reference row itself still aggregates. \texttt{claude-sonnet-4-5} exposes no \texttt{reasoning\_effort} parameter, hence the blank entry.}\label{fig:changes_without_agg}
\end{figure*}

\paragraph{Results without verdict aggregation.} Figure~\ref{fig:changes_without_agg} repeats the controlled study with aggregation turned off, calling pointwise judges once and pairwise judges once per presentation order. The reference configuration is the one from Figure~\ref{fig:ablation} and still aggregates, so each cell reflects both the configuration change and the removal of aggregation. The trends reported in \S\ref{sec:results} persist. Pairwise scores drop once aggregation is removed, so gains shrink and losses deepen. For \texttt{gpt-5.4} on \humangen, the gain from P3 drops from $+15.6$ to $+12.3$ points, while the loss from P4, already the largest in the study, deepens from $-37.0$ to $-46.8$. Pointwise judges are less sensitive to aggregation: most scores drop by a few points, and some even rise slightly.

\begin{figure*}[!htb]
    \centering
    \includegraphics[width=\textwidth]{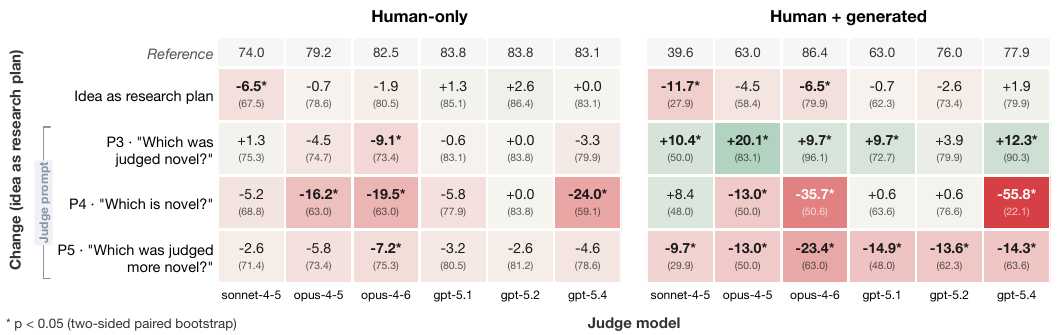}
    \caption[Prompt variants with research plans]{\textbf{Prompt variants with ideas as research plans}, pairwise strict-accuracy (ties count as failures). The first row applies the plan format alone, and rows P3--P5 combine it with the corresponding judge prompt. Scores are generally lower than with abstracts, but the effects of the prompt variants keep their direction.}\label{fig:plan_format}
\end{figure*}

\paragraph{Prompt variants on research plans.} Judges perform slightly worse when ideas are presented as research plans rather than abstracts (the \emph{Idea as research plan} row of Figure~\ref{fig:ablation_pairwise}), mostly in the \humangen setup. To check whether the plan format also changes the effect of other design choices, we rerun the pairwise prompt variants P3--P5 (which produced the largest performance changes on free-form abstracts) with every idea rewritten as a plan. Figure~\ref{fig:plan_format} reports the results against the same reference as Figure~\ref{fig:ablation_pairwise} (the reference prompt on abstracts), so each cell reflects both the format change and the prompt change. Scores are generally a bit lower than with abstracts, but the overall trends persist: P3 still improves every judge on \humangen, P5 still degrades every judge in both setups, and P4 still produces the steepest drops, down to $-55.8$ points for \texttt{gpt-5.4} on \humangen. On \humanonly, P4 also significantly degrades three judges, including \texttt{gpt-5.4} by $24$ points (about $7$ on abstracts), so its apparent robustness on human-authored ideas does not survive the format change.

\subsubsection{Different Ideation Models}\label{sub:different_ideation_models}

\begin{figure*}[!htb]
\centering
\includegraphics[width=\textwidth]{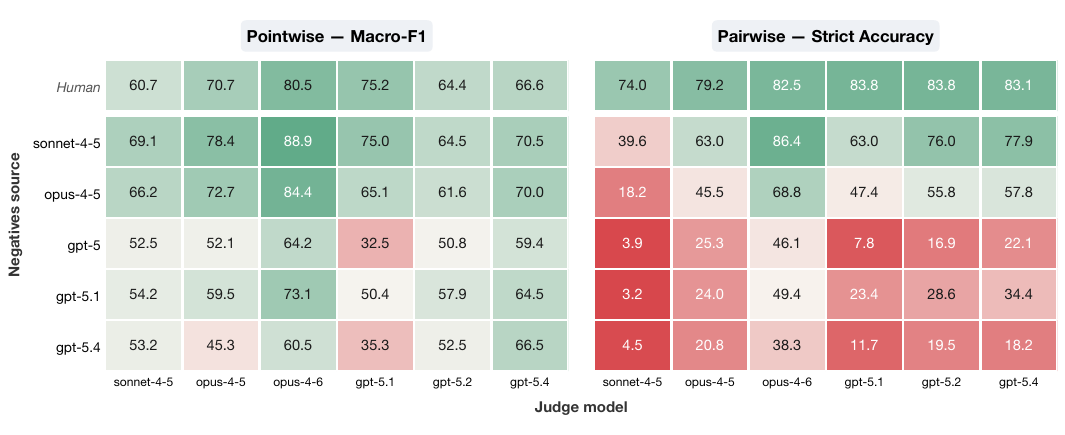}
\caption[Source of lower-novelty ideas]{\textbf{Effect of the source of lower-novelty ideas.} Judge performance when the pool of high-novelty, human-authored ideas ($\Dpos$) is held fixed and the pool of lower-novelty ideas ($\Dneg$) is regenerated by each of five ideation backbones. \humanonly, where $\Dneg$ holds human-authored ideas criticized for lacking novelty, is shown as a reference. Judge performance drops once the low-novelty side is LLM-generated, falling below chance level in some instances.}
\label{fig:negatives_source}
\end{figure*}

To examine the effect of the idea generation model, we hold $\Dpos$ fixed and regenerate $\Dneg$ with five ideation backbones (\texttt{claude-sonnet-4-5}, \texttt{claude-opus-4-5}, \texttt{gpt-5}, \texttt{gpt-5.1}, \texttt{gpt-5.4}), keeping the rest of the reference configuration (\S\ref{sub:reference_config}) unchanged. Figure~\ref{fig:negatives_source} reports judge performance under each generator, alongside the \humanonly setup as a reference point, where $\Dneg$ is human-authored.

Performance generally degrades once the negatives are LLM-generated, but pointwise and pairwise judges are not affected equally: pairwise accuracy suffers the most, falling below chance in several cases, whereas pointwise macro-F1 degrades more mildly and occasionally even improves. We also notice that the size of the drop does not necessarily track the strength of the generator: most pointwise judges rate ideas from \texttt{claude-sonnet-4-5} as no less novel than those from the stronger \texttt{claude-opus-4-5}, and using \texttt{gpt-5} ideas (whose cutoff is in 2024) degrades judges about as much as using the newer \texttt{gpt-5.4} (with a cutoff in 2025). This suggests that testing a novelty judge against human-authored ideas alone, or even against a single ideation system, is not enough: a judge that looks reliable against one type of data can invert against another.

\begin{figure}[!htb]
    \centering
    \includegraphics[width=\columnwidth]{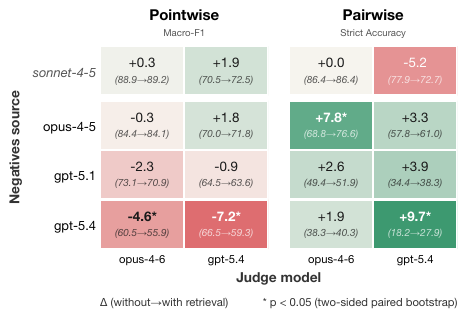}
    \caption[Retrieval across ideation backbones]{\textbf{Retrieval across ideation backbones.} Effect of giving the judge retrieved related work, with $\Dpos$ held fixed and $\Dneg$ regenerated by each of four ideation backbones (rows), for one judge per model family (columns). Retrieval yields small and inconsistent effects that do not agree in sign across the two formats, and it leaves the hardest backbones far below chance.}\label{fig:retrieval_by_generator}
\end{figure}

\paragraph{Retrieval across ideation backbones.} Figures~\ref{fig:ablation} and~\ref{fig:retrieval_faceoff} show that retrieval yields limited gains for novelty evaluation. We ask whether this still holds when $\Dneg$ is generated by different backbones. The full grid of judges $\times$ generators $\times$ design choices grows quickly, so we rerun retrieval across four ideation backbones with a single judge from each model family (\texttt{claude-opus-4-6} and \texttt{gpt-5.4}), and leave the complete cross-product to future work. Figure~\ref{fig:retrieval_by_generator} reports the results. Consistent with our earlier findings, gains from retrieval are small and inconsistent, and in some cases retrieval significantly degrades performance.

\subsubsection{Comparing Two Ideation Systems}\label{sub:ai_vs_ai}

A common use of an LLM novelty judge is to compare two ideation systems: given the same research area, which system proposes the more novel idea? We simulate this setting with two backbones from the same model family, a stronger and a weaker one: \texttt{claude-opus-4-5} vs.\ \texttt{claude-sonnet-4-5}, and \texttt{gpt-5.4} vs.\ \texttt{gpt-5.1}. We reuse the generated ideas described above, where each backbone received the same idea generation prompt (Prompt~\ref{prompt:prompted-llm-idea-gen}), seeded with the ICLR primary area of each of the $154$ pairwise \humangen instances. For each pair, we pit the idea of the stronger backbone against the idea of the weaker one, keeping the area and the presentation order, so no human-authored idea is involved.

Which of the two systems proposes more novel ideas is not known in advance: a stronger model is not necessarily a more original ideator. We therefore do not score the verdicts against an assumed label. Instead, we count \emph{verdict flips}: pairs whose verdict changes when the judge configuration changes, which requires no gold label. We focus on the prompt variants P3--P5, which produced the largest shifts in judge behavior in the main study (Figure~\ref{fig:ablation_pairwise}), and, as in the main study, compare each of them against the reference prompt. Since the three also share one lineage and differ only in the question they pose, we additionally compare P4 and P5 against P3, which isolates the effect of changing a prompt within its lineage. All other design choices follow the reference configuration (\S\ref{sub:reference_config}), including verdict aggregation. 
Table~\ref{tab:ai_vs_ai_alt} reports, per judge, the \emph{win rate} of the stronger system: the share of pairs in which the judge prefers its idea, where a tie counts as no win. We add BCa 95\% confidence intervals from a non-parametric bootstrap of $100{,}000$ resamples over the $154$ pairs.

Every prompt change flips between $13.0\%$ and $56.5\%$ of a judge's verdicts. Claude judges are the least stable ($26.0$--$56.5\%$), while GPT judges flip $13.0$--$25.3\%$ of their verdicts. Replacing P3 with P4--P5 is about as disruptive as replacing the reference prompt outright (Table~\ref{tab:ai_vs_ai_main}). These flips move the win rate of the stronger system, most of all for Claude judges: for \texttt{claude-sonnet-4-5} on the Claude pair, it ranges from $38.3\%$ under the reference prompt to $66.9\%$ under P3. GPT judges move by at most $7.1$ points. The verdict on a given pair, and the measured advantage of the stronger system, thus depend on the judge configuration.

\begin{table}[!htb]
\centering
\footnotesize
\setlength{\tabcolsep}{3pt}
\begin{tabular}{@{}l cc@{}}
\toprule
\textbf{Change} & \scriptsize \texttt{opus-4-5} vs.\ \texttt{sonnet-4-5} & \scriptsize \texttt{gpt-5.4} vs.\ \texttt{gpt-5.1} \\
\midrule
Ref.\ $\to$ P3 & 33.8 {\scriptsize (16--56)} & 27.4 {\scriptsize (16--42)} \\
Ref.\ $\to$ P4 & 26.8 {\scriptsize (15--51)} & 27.9 {\scriptsize (18--47)} \\
Ref.\ $\to$ P5 & 26.0 {\scriptsize (16--46)} & 23.9 {\scriptsize (13--34)} \\
\addlinespace[2pt]
P3 $\to$ P4 & 28.8 {\scriptsize (15--43)} & 26.4 {\scriptsize (15--37)} \\
P3 $\to$ P5 & 30.5 {\scriptsize (18--47)} & 27.4 {\scriptsize (20--38)} \\
\bottomrule
\end{tabular}
\caption[Verdict flips when comparing ideation systems]{\textbf{Judges are unstable when comparing ideation systems.} \% of pairs whose verdict changes under prompt changes (\emph{Flips}), as \texttt{mean (range)} over six judges.}
\label{tab:ai_vs_ai_main}
\end{table}

\begin{table*}[!htb]
\centering
\footnotesize
\setlength{\tabcolsep}{4pt}
\begin{tabular}{@{}l cccccc@{}}
\toprule
& son-4-5 & op-4-5 & op-4-6 & 5.1 & 5.2 & 5.4 \\
\midrule
\rowcolor{palWarmTan!18}[0pt][0pt]
\multicolumn{7}{@{}c@{}}{\textit{\texttt{opus-4-5} vs.\ \texttt{sonnet-4-5}}} \\
\quad Ref. & 38.3 {\scriptsize [30.5, 46.1]} & 59.1 {\scriptsize [51.3, 66.9]} & 59.7 {\scriptsize [51.9, 67.5]} & 61.7 {\scriptsize [53.9, 69.5]} & 64.3 {\scriptsize [56.5, 71.4]} & 65.6 {\scriptsize [57.8, 72.7]} \\
\quad P3 & 66.9 {\scriptsize [59.1, 74.0]} & 58.4 {\scriptsize [50.6, 66.2]} & 50.0 {\scriptsize [42.2, 57.8]} & 65.6 {\scriptsize [57.8, 72.7]} & 62.3 {\scriptsize [54.5, 69.5]} & 66.2 {\scriptsize [58.4, 73.4]} \\
\quad P4 & 60.4 {\scriptsize [52.6, 68.2]} & 66.9 {\scriptsize [59.1, 74.0]} & 52.6 {\scriptsize [44.8, 60.4]} & 60.4 {\scriptsize [52.6, 68.2]} & 62.3 {\scriptsize [54.5, 69.5]} & 61.7 {\scriptsize [53.9, 68.8]} \\
\quad P5 & 52.6 {\scriptsize [44.8, 60.4]} & 58.4 {\scriptsize [50.6, 66.2]} & 47.4 {\scriptsize [39.6, 55.2]} & 61.7 {\scriptsize [53.9, 69.5]} & 61.7 {\scriptsize [53.9, 69.5]} & 64.3 {\scriptsize [56.5, 71.4]} \\
\addlinespace[2pt]
\quad \textit{Spread} & 28.6 & \phantom{0}8.4 & 12.3 & \phantom{0}5.2 & \phantom{0}2.6 & \phantom{0}4.5 \\
\midrule
\rowcolor{palWarmTan!18}[0pt][0pt]
\multicolumn{7}{@{}c@{}}{\textit{\texttt{gpt-5.4} vs.\ \texttt{gpt-5.1}}} \\
\quad Ref. & 53.9 {\scriptsize [46.1, 61.7]} & 63.0 {\scriptsize [55.2, 70.1]} & 64.3 {\scriptsize [56.5, 71.4]} & 64.3 {\scriptsize [56.5, 71.4]} & 68.8 {\scriptsize [61.0, 76.0]} & 64.9 {\scriptsize [57.1, 72.1]} \\
\quad P3 & 68.8 {\scriptsize [61.0, 76.0]} & 63.0 {\scriptsize [55.2, 70.1]} & 55.8 {\scriptsize [48.1, 63.6]} & 67.5 {\scriptsize [59.7, 74.7]} & 70.8 {\scriptsize [63.0, 77.3]} & 64.3 {\scriptsize [56.5, 71.4]} \\
\quad P4 & 59.7 {\scriptsize [51.9, 67.5]} & 66.9 {\scriptsize [59.1, 74.0]} & 55.8 {\scriptsize [48.1, 63.6]} & 66.2 {\scriptsize [58.4, 73.4]} & 67.5 {\scriptsize [59.7, 74.7]} & 64.9 {\scriptsize [57.1, 72.1]} \\
\quad P5 & 56.5 {\scriptsize [48.7, 64.3]} & 69.5 {\scriptsize [61.7, 76.6]} & 63.6 {\scriptsize [55.8, 70.8]} & 60.4 {\scriptsize [52.6, 68.2]} & 65.6 {\scriptsize [57.8, 72.7]} & 63.6 {\scriptsize [55.8, 70.8]} \\
\addlinespace[2pt]
\quad \textit{Spread} & 14.9 & \phantom{0}6.5 & \phantom{0}8.4 & \phantom{0}7.1 & \phantom{0}5.2 & \phantom{0}1.3 \\
\bottomrule
\end{tabular}
\caption[Comparing two ideation systems]{\textbf{Win rate (\%) of the stronger ideation system}, per judge, over $154$ pairs, with 95\% bootstrap CIs in brackets. Each cell is the share of pairs in which the judge prefers the idea of the stronger backbone; ties count as no win. \emph{Ref.} is the reference prompt; P3--P5 share one lineage. \emph{Spread} is the gap between a judge's highest and lowest win rate across prompts. Judges are abbreviated as in Table~\ref{tab:tie_rates}.}
\label{tab:ai_vs_ai_alt}
\end{table*}

\subsection{When Retrieval Hurts}\label{sub:retrieval_qualitative}
To understand why retrieval (counterintuitively) yields limited gains (\S\ref{subsec:retrieval_limitations}), we inspect cases where it turns a correct verdict incorrect. We focus on one strong judge, \texttt{gpt-5.4}, on pointwise \humangen, where retrieval flips 22 of its correct verdicts to incorrect. Since each verdict aggregates three calls, a flip in the aggregated verdict could reflect sampling noise rather than an effect of retrieval. We therefore only consider ideas where the reference judge (no retrieval) is correct in all three calls and the retrieval judge is incorrect in all three. We sample five flips per class ($5$ from $\Dpos$ and $5$ from $\Dneg$ the reference judge classifies correctly). For each, we compare the judge's reasoning with and without retrieval. We observe a few recurring patterns (see examples for each in Table~\ref{tab:retrieval_failures}).

\paragraph{A single close precursor decides the verdict.}
This pattern underlies the $\Dpos$ flips (the reference judge correctly labels an idea novel and the retrieval judge labels it not novel). We observe that when one retrieved paper shares the idea's headline framing, the judge treats the remaining contribution as incremental, e.g., a larger benchmark as ``primarily a scale/resource extension'', or a new failure mode as ``mainly a setting shift'' (even when this part of the idea is praised by reviewers).

\paragraph{Component-wise matching.}
This pattern underlies the $\Dpos$ flips (the reference judge correctly labels an idea novel and the retrieval judge labels it not novel). The retrieval judge maps each component of the idea to a different retrieved paper and concludes that the whole is ``a fairly direct synthesis'' (even when human reviewers explicitly describe the idea as a novel combination). Notably, without retrieval, the judge often recognizes the same components as known (``The core ingredients are not individually new'') yet still credits their combination as novel. 

\paragraph{Retrieved work displaces prior knowledge.}
This pattern underlies the $\Dneg$ flips (the reference judge correctly labels a generated idea not novel and the retrieval judge labels it novel). Without retrieval, the judge dismisses the generated ideas as not novel based on its internal knowledge. With retrieval, it checks each idea against the five retrieved papers, finds none that covers it fully, and concludes it is novel (``does not appear in the provided related work'', ``None of the cited papers ...''). Our retrieval prompt defines novelty ``in relation to the provided related work'' (Prompt~\ref{prompt:diff-retrieval}), which likely encourages this over-reliance on retrieved papers.

\begin{table*}[t]
\centering
\small
\begin{tabularx}{\textwidth}{@{}X@{}}
\toprule
\textbf{Idea} (\Dpos): ``...Multi-Agent System (MAS) powered by Visual Language Models ... suffers from a novel failure term, multi-agent visual hallucination snowballing, where hallucinations are seeded in a single agent and amplified by following ones... we propose ViF, a lightweight, model-agnostic mitigation paradigm...'' \\
\textbf{Reviewers}: ``The authors introduce a new task setting for MAS hallucination and create a multi-turn evaluation metric which could be beneficial for future researchers.'' \\
\textbf{Without retrieval}: ``...although several ingredients have precedents, the problem setting plus the specific diagnosis-and-mitigation package appears substantively original...'' $\rightarrow$ novel \faCheck \\
\textbf{Retrieved}: \textit{Investigating and Mitigating the Multimodal Hallucination Snowballing in Large Vision-Language Models} (2024)... \\
\textbf{With retrieval}: ``...although the MAS-specific framing is somewhat new, the main concept, analysis direction, and mitigation ingredients are all strongly anticipated by the provided related work...'' $\rightarrow$ not novel \faTimes \\
\midrule
\textbf{Idea} (\Dpos): ``...we revisit Sobolev integral probability metrics (IPM) on trees to obtain a practical generalization of TSW...'' \\
\textbf{Reviewers}: ``The idea of combining tree-sliced Wasserstein with the Sobolev IPM to handle the general $p$-Wasserstein distance is novel'' \\
\textbf{Without retrieval}: ``The core ingredients are not individually new... However, the specific contribution described here appears to be a genuinely new synthesis'' $\rightarrow$ novel \faCheck \\
\textbf{Retrieved}: [1] \textit{Sobolev Transport: A Scalable Metric for Probability Measures with Graph Metrics} (2022); ... ; [4] \textit{Tree-Sliced Variants of Wasserstein Distances} (2019) \\
\textbf{With retrieval}: ``a direct combination of existing ingredients. [1] already introduces the tree-sliced paradigm... [4] already introduces a closed-form Sobolev transport / Sobolev IPM on graph metric spaces.... Therefore, defining “Tree-Sliced Sobolev” by averaging a Sobolev/tree discrepancy over random trees is a fairly direct specialization/composition of [1] and [4], rather than a new framework.'' $\rightarrow$ not novel \faTimes \\
\midrule
\textbf{Idea} (\Dneg, generated): ``We introduce ChemoSense, a multimodal framework that learns computational representations of taste and smell...by leveraging their correlations with vision, language, audio, and molecular structure'' \\
\textbf{Retrieved}: \textit{SMILES to Smell} (2021); \textit{Recipe1M+} (2021); \textit{DeepNose} (2018); \textit{VirtualTaste} (2021); a recipe-retrieval model (2021) \\
\textbf{Without retrieval}: ``this looks more like a broad composition of known components than a clearly novel framework, task, or resource.'' $\rightarrow$ not novel \faCheck \\
\textbf{With retrieval}: ``...the idea is not novel at the level of basic method motifs. However the proposal does appear novel in its overall problem formulation and synthesis relative to the provided related work. None of the cited papers explicitly ... align chemical compounds with culinary multimodal data ... in one shared embedding space...'' $\rightarrow$ novel \faTimes \\
\bottomrule
\end{tabularx}
\caption[Retrieval hurting \texttt{gpt-5.4}]{\textbf{Examples of retrieval hurting \texttt{gpt-5.4}} (\humangen, pointwise). In the first row, a single close precursor outweighs the contribution the reviewers valued. In the second, the judge matches each component to a different retrieved paper and dismisses the combination the reviewers called novel. In the third, the retrieved papers override the judge's own recognition that the idea is a broad composition of known components: since no single retrieved paper covers the idea in full, the judge concludes it is novel.}
\label{tab:retrieval_failures}
\end{table*}

\end{document}